\documentclass{ecai} 

\usepackage{latexsym}
\usepackage{amssymb}
\usepackage{amsmath}
\usepackage{amsthm}
\usepackage{booktabs}
\usepackage{enumitem}
\usepackage{graphicx}
\usepackage{color}

\usepackage{todonotes}      

\usepackage{booktabs}       
\usepackage{amsthm}
\usepackage{amsmath}
\usepackage{dsfont}
\usepackage{subcaption}
\usepackage{float}
\usepackage[hidelinks]{hyperref}

\usepackage{siunitx}

\usepackage{soul,xcolor}
\setul{-2ex}{2ex}

\usepackage{dblfloatfix}

\newcommand{\BibTeX}{B\kern-.05em{\sc i\kern-.025em b}\kern-.08em\TeX}
\DeclareMathOperator*{\fBm}{fBm}
\DeclareMathOperator*{\FBM}{FBM}
\DeclareMathOperator*{\fOU}{fOU}
\DeclareMathOperator*{\FOU}{FOU}
\DeclareMathOperator*{\ARFIMA}{ARFIMA}
\DeclareMathOperator*{\Cov}{Cov}

\DeclareMathOperator*{\Var}{Var}
\DeclareMathOperator*{\Exp}{Exp}

\begin{document}


\begin{frontmatter}


\paperid{1677} 


\title{Parameter Estimation of Long Memory\\ 
Stochastic Processes with Deep Neural Networks}


\author[A]{\fnms{Bálint}~\snm{Csanády}\thanks{Corresponding authors: csbalint@protonmail.ch, andras.lukacs@ttk.elte.hu}}
\author[A,B]{\fnms{Lóránt}~\snm{Nagy}}
\author[A]{\fnms{Dániel}~\snm{Boros}}
\author[A,B]{\fnms{Iván}~\snm{Ivkovic}}
\author[A]{\fnms{Dávid}~\snm{Kovács}}
\author[A]{\fnms{Dalma}~\snm{Tóth-Lakits}}
\author[A,C]{\fnms{László}~\snm{Márkus}}
\author[A,\textasteriskcentered]{\fnms{András}~\snm{Lukács}} 

\address[A]{Institute of Mathematics, ELTE Eötvös Loránd University, Budapest, Hungary}
\address[B]{Alfréd Rényi Institute of Mathematics, Budapest, Hungary}
\address[C]{Department of Statistics, University of Connecticut, Connecticut, USA}


\begin{abstract}
    We present a purely deep neural network-based approach for estimating long memory parameters of time series models that incorporate the phenomenon of long-range dependence.
    Parameters, such as the Hurst exponent, are critical in characterizing the long-range dependence, roughness, and self-similarity of stochastic processes.
    The accurate and fast estimation of these parameters holds significant importance across various scientific disciplines, including finance, physics, and engineering.
    We harnessed efficient process generators to provide high-quality synthetic training data, enabling the training of scale-invariant 1D Convolutional Neural Networks (CNNs) and Long Short-Term Memory (LSTM) models.
    Our neural models outperform conventional statistical methods, even those augmented with neural networks. 
    The precision, speed, consistency, and robustness of our estimators are demonstrated through experiments involving fractional Brownian motion (fBm), the Autoregressive Fractionally Integrated Moving Average (ARFIMA) process, and the fractional Ornstein-Uhlenbeck (fOU) process.
    We believe that our work will inspire further research in the field of stochastic process modeling and parameter estimation using deep learning techniques.
\end{abstract}

\end{frontmatter}


\section{Introduction}
    Long-range dependence, or long memory, plays a critical role in the scientific modeling of natural and industrial phenomena.
    On the one hand, in the field of natural sciences, long memory finds application in diverse areas such as climate change \citep{Yuan.etal2022,franzke2015dynamical}, hydrology \citep{hurst1956problem}, detection of epilepsy \citep{acharya2012application}, DNA sequencing \citep{Lopes2006}, data networks \citep{Willinger2001LongRangeDA}, and in cybersecurity through anomaly detection \citep{li2006change}.
    On the other hand, research on long memory implies achievements in financial mathematics, see for example \citep{qian2004hurst, eisler2006size} and \citep{Baillie1996} for the application of long memory in volatility modeling.
    The ubiquity of long memory across time series data has drawn considerable attention to models capable of capturing this phenomenon.
    In most stochastic models, the impact of past events on future events decays rapidly, making the effect of observations from the distant past negligible in terms of forecasting ability.
    If long-range dependence is present in a system, predictions concerning the future require information from the complete history of the process, in contrast to Markovian environments,
    where the most recent events already contain all the necessary information for an optimal forecast.
    Modern literature associates long-range dependence with a slow hyperbolic decay, namely when autocorrelations $\rho_t$ vanish with a rate that is comparable to $t^{-\alpha}$ for some $\alpha \in (0,1)$ \citep{Giraitis2012}.  
    The range $\alpha \in (1,2)$ is often referred to as anti-persistence.
    In broader terms, ``long memory'' can be used to mean that autocorrelations have a longer decay than exponential.
    
    When one models data with long memory, it is a crucial task to estimate model parameters, and classical inference methods are often not applicable in the case of long memory processes.
    We focus our attention on three stochastic processes that are frequently utilized in modern applied mathematics: the fractional Brownian motion (fBm), the Autoregressive Fractionally Integrated Moving Average (ARFIMA), and the fractional Ornstein-Uhlenbeck (fOU) process.
    In the case of fBm and fOU, we concentrate on estimating the Hurst parameter.
    The Hurst exponent controls the roughness, self-similarity, and long-range dependence of fractional Brownian motion paths, and this way also influences the characteristics of derivative processes such as the fractional Ornstein-Uhlenbeck process.
    Regarding ARFIMA models, our target is the differencing parameter $d$, governing the decay of autocovariances, and thus, the decay of memory in the system.
    
  \subsection{Main results}
    We propose the adoption of well-known neural network architectures as versatile tools for estimating the parameters that characterize long memory.
    A key aspect of our approach is that efficient process generators provide the opportunity to train the models on large amounts of data.
    This facilitates better performance compared to traditional statistical estimation methods, even those supplemented with neural networks.
    Inference is efficient in terms of speed, even for samples of long sequences, making our approach valuable for practical applications.
    We can also guarantee various invariances such as shift, drift, and scale-invariance.
    A further advantage of our method is that it performs consistently across the entire $(0,1)$ range of the Hurst parameter.
    We found that the proposed neural network estimator improves in accuracy when trained on longer sequences, a consistency that persists even during inference on sequences of lengths different from those used in training.
    The above virtues are supported by the measurements presented in the paper.
    Our code is publicly available on GitHub.\footnote{\href{https://github.com/aielte-research/LMSParEst}{https://github.com/aielte-research/LMSParEst}}

  \subsection{Related work}
    In recent years, a number of studies have emerged that utilize neural networks to estimate the Hurst parameter of fBm.
    Some of these works apply MLPs which, due to their fixed-size input requirement, present one of the following alternatives: inference can either be performed only on a fixed-length series \citep{ledesma2011hurst,han2020deciphering}, or on a set of process specific statistical measures enabling the fixed-size input to the neural networks \citep{kirichenko2022wavelet,mukherjee2023hurst}.
    A more general approach is the signature-based method described in \citep{kidger2019deep}, 
    which can also be used to estimate fBm Hurst, where the extracted statistical descriptors are processed by an LSTM.
    In the case of these methods, the hybrid application of statistical descriptors and neural networks brings less improvement compared to our purely neural network solutions.
    This is reflected in the comparison to several classical estimation methods.
    Another shortcoming in recently published methods is that they do not address the possible limitations caused by scaled inputs.
    Regarding the ARFIMA and fOU processes, so far, we could not find neural network based parameter estimators in the literature.
    We note that our estimator's architecture is close to the work by \citet{wang2022neural} which was used for the non-fractional Ornstein-Uhlenbeck process.
    A more complex architecture in subsequent work by \citet{feng2023deep} estimates the Hurst parameter in fBm.
    In the case of the fOU process, we compared the neural network-based Hurst estimates with a quadratic variation estimator \citep{brouste2011}, and our method presented much higher accuracy.
    For the ARFIMA process Whitte's method \citep{Whittle1951} yielded comparable performance to our approach.
    
    The success of the utilized networks (Section \ref{subsec:NNarchitecture}) largely 
    stems from a large volume of high-quality training data, manifested with the software that was built around the framework of the so-called isonormal processes (see \cite{Nualart:2018} for the mathematical background, and Section \ref{subsec:generating} on the implementation).
    The underlying path-generating methodology includes the circulant embedding of covariance matrices and the utilization of fast Fourier transform (FFT).

\section{Processes}\label{sec:background}
    This section provides a summary of the most important properties of the stochastic processes discussed  in this paper.
    
  \subsection{Fractional Brownian motion}
    The fractional Brownian motion $\fBm(H):=\big(B_t^H\big)_{t \geq 0}$, $H \in (0, 1)$ is a continuous centered Gaussian process  with covariance function
    \begin{equation*}\label{fBm_eq}
         \displaystyle\Cov\!\left({B_t^H,B_s^H}\right) = \frac{1}{2}\!\left({\lvert{t}\rvert^{2H} + \lvert{s}\rvert^{2H} - \lvert{t-s}\rvert^{2H}}\right).
    \end{equation*}
    Here, $H$ is called the Hurst exponent of the process.
    Let $\FBM(H, n, S, T):\:S<T$  denote the distribution of the $(T-S)/n$-equidistant realizations of $\fBm(H)$ on the time interval $[S,\,T]$.
    If $\Delta \FBM(H, n, S, T)$ denotes the sequence of increments, it can be shown that $\Delta \FBM(H, n, S, T) \sim \lambda(H,T-S)\Delta \FBM(H, n, 0, 1)$ where $\lambda(H,T-S)$ is a scalar.
    If we want to estimate $H$ from $\FBM(H, n, S, T)$ we might want to consider a shift invariant neural network on the increments, since then it will be sufficient to train it only on $\FBM(H, n, 0, 1)$.
    We might also consider the scaled and drifted fBm process $\fBm(H,\sigma,\mu):=\big(\sigma B_t^H + \mu t \big)_{t \geq 0}$, with $H \in (0, 1), \sigma > 0, \mu \in \mathbb{R}$, which is the fractional counterpart of the so called Bachelier model \citep{musiela2006martingale}.
    When the network is also drift invariant, it is still sufficient to train the network on realizations of $\FBM(H, n, 0, 1)$ to yield an estimator for the parameter $H$ of $\fBm(H,\sigma,\mu)$.
        
  \subsection{Autoregressive Fractionally Integrated Moving Average}
    A covariance stationary sequence of random variables $\zeta_j, \ j\in \mathbb{Z}$ is said to form white-noise if $E\zeta_0=0$, $\gamma(0)=E\zeta_0^2<\infty$, and $\gamma(k) = 0$ for all $k \in \mathbb{Z}, \ k \neq 0$.
    For $d > -1$ the fractional difference operator is defined as 
    \begin{equation*}
        \nabla^d = \sum_{k=0}^{\infty} {d \choose k}(-B)^{-k},
    \end{equation*}
    where $B$ is the backward shift operator, that is $BX_j = X_{j-1}$.
    We define the $\ARFIMA(0,d,0)$ process for $d \in (-1/2,1/2)$ as the solution of the difference equation
    \begin{equation}\label{arfima_eq}
        \nabla^d X_j = \zeta_j,
    \end{equation}
    where $\zeta_j, \ j\in \mathbb{Z}$ is a white-noise sequence.
    It is known that when $\zeta_j, \ j\in \mathbb{Z}$ is ergodic (e.g. in the sense of Definition 2.8.4 in \cite{gikhman2004theory}), and $d \neq 0$, there exists a unique stationary solution to Equation~(\ref{arfima_eq}) -- see Theorem 7.2.2 in \cite{Giraitis2012}.

  \subsection{The fractional Ornstein-Uhlenbeck process} \label{sec:fOU_bg}
    Let $H \in (0, 1), \alpha, \sigma > 0,  \eta, \mu \in \mathbb{R}$.
    The fractional Ornstein-Uhlenbeck process $(Y_t)_{t \geq 0}$ is the solution of the following stochastic differential equation:
    \begin{flalign*}
    \label{fou}
    dY_t &= -\alpha (Y_t - \mu) \, dt + \sigma \, dB^H_t \\
    Y_0 &= \eta.
    \end{flalign*}
    Let $\fOU(\eta, H, \alpha, \mu, \sigma)$ denote the distribution of this process on the Borel $\sigma$-algebra of continuous functions.
    Note that $\mu$ and $\sigma$ are scaling and shifting parameters.
    If $Y \sim \fOU\big((\eta - \mu) / \sigma, H, \alpha, 0, 1)$, then $\sigma Y + \mu \sim \fOU(\eta, H, \alpha, \mu, \sigma)$.
     This means that if we can guarantee the scale and shift invariance of the network, it will be sufficient to train a $H$-estimator on realizations from $\fOU\big(\eta, H, \alpha, 0, 1)$ to cover the distribution on $\fOU(\eta, H, \alpha, \mu, \sigma)$.

\section{Baseline estimators}\label{sec:baselines_main}
    To provide baseline comparisons to our neural network 
    results, we considered the following statistical estimators.

  \subsection{Rescaled range analysis}
    The term and concept of rescaled range analysis stem from multiple works of Harold E. Hurst; see, for example, \citep{hurst1956problem} for a study in hydrology, and \citep{graves2017brief} for a historical account on the methodology.
    The statistics $R/S$, defined below, is the rescaled and mean adjusted range of the progressive sum of a sequence of random variables, more precisely, given $Z_1, Z_2, ...$, for a positive integer $n$ consider the statistics
    \begin{equation}\label{RdashS}
    R/S(n) = \frac{\max\limits_{1\leq k \leq n}\left\{ X_k - \frac{k}{n}X_n\right\} - \min\limits_{1\leq k \leq n}\left\{ X_k - \frac{k}{n}X_n\right\}}{\sqrt{\frac{1}{n} \sum\limits_{k=1}^n \left(  Z_k - \frac{1}{n} X_n \right)^2}},
    \end{equation}
    where $X_k = \sum_{i=1}^k Z_i$.
    The analysis is done via assuming an asymptotics for the statistics in Equation~(\ref{RdashS}), namely we postulate that on the long run, it holds that $R/S(n) \approx cn^{h}$, where $c$ is an unknown constant and $h$ is the parameter we are looking for.
    Utilizing a one parameter log-log regression on the above formula, that is, using the relation $\log(R/S(n)) = \log(c) + h\log(n)$, one can estimate $h$.
    
    Turning to fractional Brownian motion, it is shown in \citep{taqqu1995estimators}, that its increment process, fractional Gaussian noise has the property $R/S(n) \approx c_0 n^H$,
    where $H$ is the Hurst parameter of the underlying fractional process, and $c_0$ is some positive constant: yielding a numerical method for the estimation of $H$.
    
    A known limitation of this methodology, when the underlying process is a fractional Brownian motion, is that using the statistics in (\ref{RdashS}) produces inferred values that are lower when the true value of $H$ is in the long memory range, and substantially higher values when the time series shows heavy anti-persistence. 
    A possible mitigation of this is to introduce a correction that calibrates the method to fit fractional Brownian motion data and use the corrected estimator as a baseline.

  \subsection{Variation Estimators}
    For a stationary increment Gaussian process, consider the lag $t$ second order statistics defined by
    \begin{equation}\label{roka}
        \gamma_p(t) = \frac{1}{2}\mathbf{E}|X_{t} - X_0|^p.
    \end{equation}
    This, when setting $p=2$, is also called the variogram of order $2$. Assume that the behavior of $\gamma_{2}(t)$ at the coordinate origin (that is the asymptotics when $t \to 0$) is like that of $|t|^{\alpha}$, for some fractal index $\alpha \in (0,2]$.
    Then, the relationship between the Hausdorff dimension and the fractal index, $$D = 2 - \frac{\alpha}{2}$$ holds -- assuming a topological dimension of $1$ -- and performing a log-log regression yields an estimate for $D$.
    
    However, the relationship between the Hausdorff dimension and the fractal index appears to be more robust and inclusive when we choose $p=1$. In this case, when similar asymptotics holds for $\gamma_1(t)$, estimators presented e.g. in \citep{gneiting2012estimators} give statistically efficient procedures to determine the Hausdorff dimension of Gaussian processes.
    In our work, the Hurst parameter of fractional processes is then estimated using the relationship $D = 2 - H$. For detailed notes on how $\gamma$ is approximated and the optimality of the choice $p=1$ we refer to \citep{gneiting2012estimators}.

  \subsection{Higuchi's method}
    Higuchi's method \citep{higuchi} relies on the computation of the fractal dimension by a one dimensional box counting.
    For a sliding box size $b \in \mathbb{N}$ and a starting point $i \in \mathbb{N}, i \leq b$, consider
    \begin{flalign*}
    L_b(i) = \displaystyle \frac{1}{{\left[\frac{n-i}{b}\right]}}\sum^{\left[\frac{n-i}{b}\right]}_{k=1}\left|{X_{i + kb}-X_{i + (k-1)b}}\right|\,.
    \end{flalign*}
    Then let $L_b := \frac{1}{b} \sum_{i = 1}^{b} L_b(i)$.
    If $X \sim \fBm(H, 0, \sigma)$ then $E(L_b) = c b^{H}$ holds.
    Thus, the slope coefficient of the linear regression $\log(L_b) \sim \log(b)$ yields an estimate for $H$. 

  \subsection{Whittle's method}
    The likelihood procedure dubbed as Whittle's method, see e.g. \citep{Whittle1951}, is based on approximating the Gaussian log-likelihood of a sample of random variables $X = (X_1,...,X_n)$, where the underlying process is stationary and Gaussian.
    We give the details in the case when we wish to estimate the Hurst parameter of fractional Brownian motion with Hurst parameter $H \in (0,1)$, and we apply Whittle's method on its increments.
    Denoting with $\Gamma_H$ the covariance matrix corresponding to the vector $X$, the likelihood of the sample with respect to $H$ can be written as $L(X) = (2 \pi )^{-n/2} \left| \Gamma_H \right|^{-1/2} e^{-\frac{1}{2}X^T \Gamma_H^{-1} X},$ where $|\Gamma_H|$ and $\Gamma_H^{-1}$ denotes the determinant and the inverse of the matrix $\Gamma_H$ respectively, and $X^T$ denotes the transpose of the vector $X$.
    To speed up the procedure, instead of numerical computations, an approximation can be introduced, see e.g. \citep{Beran2017}, and the Hurst parameter $H$ can be approximated by minimizing the quantity
    \begin{equation}\label{the_integral_likelihood}
    Q(H) = \int\limits_{-\pi}^{\pi} \frac{I(\lambda)}{f_H(\lambda)} d\lambda,
    \end{equation}
    where $I(\lambda)$ is the periodogram, an unbiased estimator of the spectral density $f_H$,  defined as $I(\lambda)=\sum_{j= -(n-1)}^{n-1}\hat\gamma(j)e^{\mathrm{i}j\lambda}$, with the complex imaginary unit $\mathrm{i}$, and where the sample autocovariance $\hat\gamma(j)$, using the sample average $\bar X = \frac{1}{n}\sum_{k=1}^n X_k$, is $\hat \gamma(j) = \sum_{k=0}^{n-|j|-1} \left( X_{k} - \bar X \right) \left( X_{k+|j|} - \bar X \right)$.
    The quantity in Equation~(\ref{the_integral_likelihood}) is usually approximated with the sum $\tilde{Q}(H) = \sum_{k=1}^{\lfloor n/2 \rfloor}\frac{I(\lambda_k)}{f_H(\lambda_k)}$, with  $\lambda_k = \frac{2 \pi k}{n}$, to obtain an asymptotically correct estimate $\hat H$ of the Hurst parameter $H$.

  \subsection{The case of ARFIMA and fOU} \label{subsec:arfima}
    According to Theorem 7.2.1 in \cite{Giraitis2012}, for the autocovariance of an ARFIMA process, we have $\gamma(k) \approx c_d k^{2d-1}$.
    Thus, in terms of the decay of autocovariance and memory properties (see Definition 3.1.2 in \cite{Giraitis2012}), the $\mbox{ARFIMA}(0,d,0)$ process corresponds to a fractional noise with Hurst parameter $H=d + 1/2$. Also, an ARFIMA process, in an asymptotic sense, has similar spectral properties to that of fractional Brownian motion incremets.
    On one hand this means, that an ARFIMA process offers a potential way to test estimators calibrated to fractional Brownian motion.
    On the other hand, it is reasonable to apply the above baseline Hurst parameter estimators for estimating the parameter $d$ of $\mbox{ARFIMA}(0,d,0)$.
    
    To estimate the Hurst parameter of a fractional Ornstein-Uhlebeck process, \citet{brouste2011} provide a statistical method (QGV) that is based on building an estimator that compares generalized quadratic variations corresponding to certain filtered versions of the input samples.
    The method presents consistent and asymptotically Gaussian estimators, and can be considered a state of the art analytical tool regarding Hurst parameter inference on fOU processes.

\section{Methods}
  \subsection{Training paradigm}\label{sec:training_paradigm}
    In contrast to a situation characterized by a limited amount of data, we can leverage synthetic data generators to train our neural network models on a virtually infinite dataset.
    The loss computed on the most recent training batches simultaneously serves as a validation loss, as each batch comprises entirely new synthetic data.
    This setup prevents overfitting in the traditional sense.
    The only potential issue associated with this training paradigm is the quality of the process generator itself.
    If the generator fails to approximate the target distribution effectively, there is a potential risk of ``overfitting'' on generated distribution and not generalizing. 
    Thus, high-quality process generators are essential.
    
    Our setup to obtain parameter estimators by utilizing generators for given families of stochastic processes is as follows:
    Let $\Theta$ be the set of the possible parameters and $P$ be the prior distribution on $\Theta$.
    For a fixed $a\in \Theta$, the generator $G^a$ denotes an algorithm that generates sample paths of a stochastic process, where the sample paths are distributed according to the process distribution $Q_a$.
    This is deterministic in the sense that every iteration returns a sample path.
    The generated sample path can also be viewed as a random object by introducing randomness into the parameter $a$, through setting $a = \vartheta$, where $\vartheta$ is a random variable distributed according to some law $P$.
    Denote the compound generator by $G^{(\vartheta)}$.
    Now, suppose we have $G^{(\vartheta)}$ as input and we would like to estimate $\vartheta$.
    Formally, an optimal $M$ estimator would minimize the MSE $E[M(G^{(\vartheta)}) - \vartheta]^2$.
    By having independent realizations of series from $G^{(\vartheta)}$, we can consider the training set $T$.
    Training a proper neural network $\mathcal{M}$ on $T$ with the squared loss function would be a heuristic attempt to obtain the above $M$ estimator.    
    We may assume that $Q$ is only parameterized by the target $a$ without loss of generality.
    If $Q$ is parameterized by other parameters besides $a$, then these can be randomized obtaining a redefined $Q_{a}$ mixed distribution.
    
  \subsection{Generating fractional processes}\label{subsec:generating}
    To generate the fractional processes fBM and fOU, we employed a circular matrix embedding method belonging to the Davies-Harte procedure family \citep{davies1987tests}.
    While the original Davies-Harte method is available in existing Python packages \citep{pypifbm}, our implementation utilizes Kroese's method \citep{kroese2014spatial}, which we adapted specifically to leverage efficient tools within the Python ecosystem for sequence generation.
    This approach also allows for storing the covariance structure, eliminating the need for repeated computations when generating multiple process realizations.
    Furthermore, it is possible to implement an optimized variant of the conventional Cholesky method, providing similar acceleration for large-scale data generation as offered by existing state-of-the-art solutions.

  \subsection{Neural network architecture}\label{subsec:NNarchitecture}
    The neural network architecture $\mathcal{M}$ is designed for regression on a sequential input, and consists of three main components:
    an optional standardizing layer to ensure invariances, a sequence transformation, and a regression head.
    There are three kinds of invariances that we might require from $\mathcal{M}$: shift, scale, and drift invariance.
    In order to make an fBm Hurst-estimator which works well in practice, we want to rely on all three of the above invariances.
    Shift invariance can be obtained by transforming the input sequence to the sequence of its increments.
    Differentiating the input this way also turns drift invariance to shift invariance.
    By performing a standardization on the sequence of increments we can ensure all three invariances.
    Such a standardizing step can also be considered as a preprocessing layer to the network, applying the transformation $x \mapsto (x - \overline{x})/{\hat{\sigma}(x)}$ to each sequence $x$ in the input batch separately, where $\hat{\sigma}(x)$ is the empirical standard deviation over the sequence $x$,
    and $\overline{x}$ is the arithmetic mean over $x$.
    Given the ergodicity properties of the underlying processes, the term ``standardizing'' remains adequate even with co-dependent data points as inputs.
    We considered two alternatives for sequence transformation.
    In $\mathcal{M}_{\text{conv}}$ the transformation is realized by a 1D CNN \citep{conv1d_theory}, and  in $\mathcal{M}_{\text{LSTM}}$, it is achieved by an LSTM \citep{lstm_theory}.
    The regression head first takes the transformed sequence and applies global average pooling resulting in an average feature vector which has a fixed dimension (the same as the output feature size in the sequence transformation).
    The head then obtains the final scalar output from the average feature vector using an MLP \citep{mlp_theory}.
    
    We found that unless it is severely underparameterized, the specific hyperparameter configuration does not have a significant effect the performance of $\mathcal{M}$.
    Generally $\mathcal{M}_{\text{LSTM}}$ achieves a somewhat smaller loss than $\mathcal{M}_{\text{conv}}$, while $\mathcal{M}_{\text{conv}}$ is computationally faster than $\mathcal{M}_{\text{LSTM}}$.
    Slight differences can arise in the speed of convergence, but these are not relevant due to the unlimited data and fast generators.
    The following are the hyperparameters that we used in the experiments below.

    We implement $\mathcal{M}_{\text{conv}}$ using a 1D convolutional network with 6 layers.
    The input has 1 channel, and the convolution layers have output channels sizes of 64, 64, 128, 128, 128, and 128.
    Every layer has stride 1, kernel size 4 and no padding.
    The activation function is PReLU after each layer.
    Our $\mathcal{M}_{\text{LSTM}}$ consists of an unidirectional LSTM with 2 layers, its input dimension is 1 and the dimension of its inner representation is 128.
    In both models, the regression head uses an MLP of 3 layers (output dimensions of 128, 64 and 1), with PReLU activation function between the first two layers.
    Unless stated otherwise the models also include a standardizing layer to ensure shift and scale invariance, and when drift invariance necessary, the model also contains a finite differentiation step.
    AdamW optimization on the MSE loss function was used for training the models \citep{adamw}.
    The learning rate was set to $10^{-4}$ and the train 
    batch size to 32.

  \subsection{Technical details}
    The process generators were implemented in Python, using \texttt{NumPy} \citep{numpy} and \texttt{SciPy} \citep{scipy}.
    We imported Higuchi's method from the package \texttt{AntroPy} \citep{antropy},
    and the R/S method from the package \texttt{hurst} \citep{hurst_package}.
    We generated the ARFIMA trajectories using the \texttt{arfima} package \citep{py_arfima}.
    The framework responsible for the training process was implemented in \texttt{PyTorch} \citep{pytorch}, where every neural module we relied on was readily available.
    The models were trained on Nvidia RTX 2080 Ti graphics cards, and we managed our experiments using the experiment tracker \texttt{neptune.ai} \citep{neptuneai2022}.
    Training took approximately one GPU hour to one GPU day per model, depending on the type of process, the applied architecture, and on the length of sequences used for training.
    A shorter training time can mostly be expected from an optimized sequence generation.

\section{Experiments} \label{experiments}
  \subsection{Metrics}
    In addition to the standard MSE loss, we use the following two metrics. 
    For a $H$-estimator $M$, let $b_{\varepsilon}(x) = m_{\varepsilon}(x) - x$ be the empirical bias function of radius $\varepsilon$,
    where $m_{\varepsilon}(x)$ is the average of inferred values 
    the estimator $M$ produces for the input sequences with $H\in[x - \varepsilon, x + \varepsilon]$.
    Similarly, let the function $\sigma_{\varepsilon}(x)$ be defined as the empirical standard deviation of the values inferred by $M$ for inputs inside the sliding window of radius $\varepsilon$.
    
    Let us denote the approximate absolute area under the Hurst--bias curve by $\hat{b}_{\varepsilon} := \varepsilon \sum_{j = 0}^{[1/\varepsilon]}|b_{\varepsilon}(\varepsilon j)|,$ and the approximate area under the Hurst--$\sigma$ curve by $\hat{\sigma}_{\varepsilon} := \varepsilon \sum_{j = 0}^{[1/\varepsilon]}\sigma_{\varepsilon}(\varepsilon j)$.
    We use these two metrics in addition to MSE, as they highlight different aspects of the estimators performance.

    \begin{figure*}[!ht]
     \centering
     \hspace*{-0.3mm}
     \begin{subfigure}[b]{.465\textwidth}
        \centering
        \includegraphics[width=0.985\textwidth]{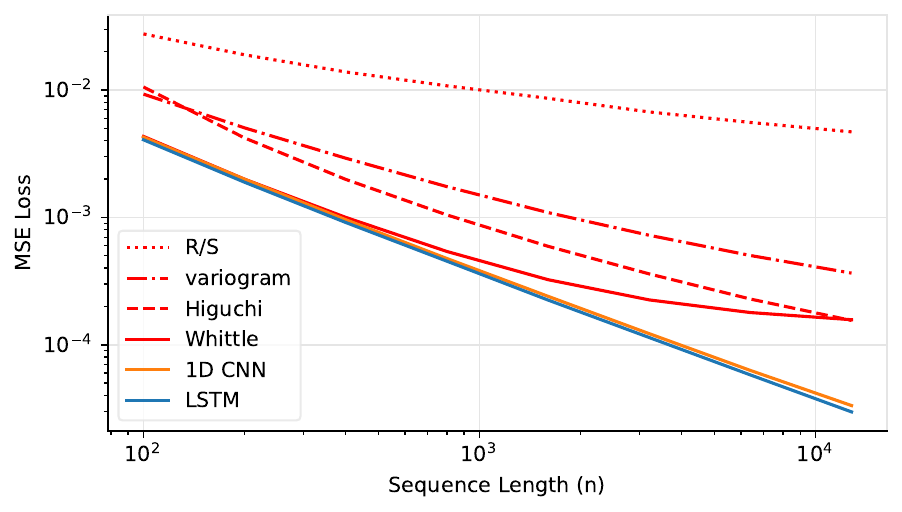}
        \subcaption{Comparison of baseline estimators and neural models.}
        \label{fig:fBm-MSE_a}
     \end{subfigure}
     \hfill
     \begin{subfigure}[b]{.465\textwidth}
        \centering
        \includegraphics[width=0.985\textwidth]{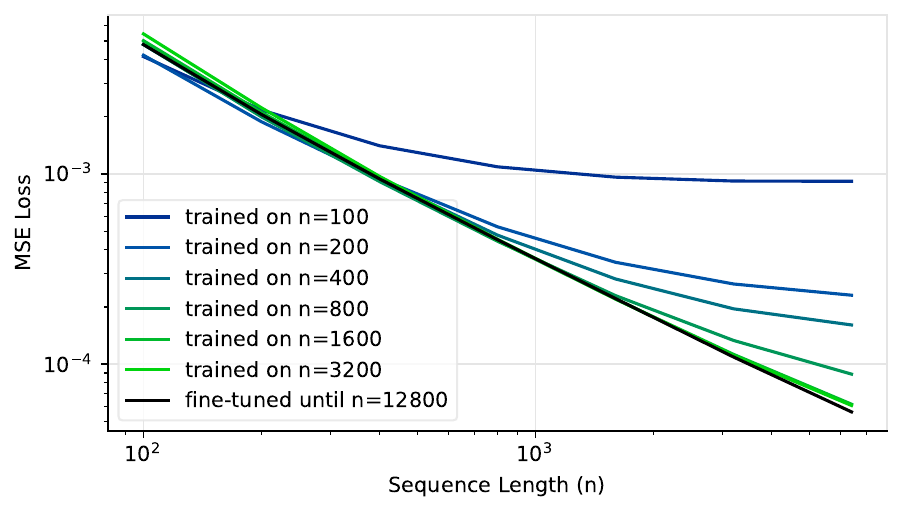}
        \subcaption{Empirical consistency of different LSTM models.}
        \label{fig:fBm-MSE_b}
     \end{subfigure}
     \hspace*{4.8mm}
     \vskip 3mm
     \caption{MSE losses of different fBm Hurst-estimators by sequence length on a log-log scale.}
    \label{fig:fBm-MSE}
\end{figure*}    
    \begin{figure*}[!ht]
     \vskip 0.9mm
     \centering
     \begin{subfigure}[b]{.465\textwidth}
        \centering
        \includegraphics[width=\textwidth]{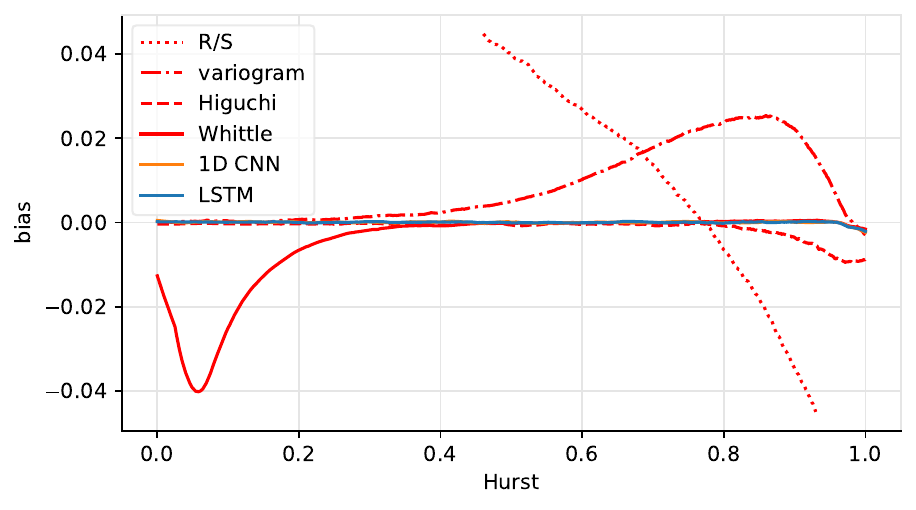}
        \subcaption{Empirical bias $b_{0.025}$.}
     \end{subfigure}
     \hfill
     \begin{subfigure}[b]{.465\textwidth}
        \centering
        \includegraphics[width=\textwidth]{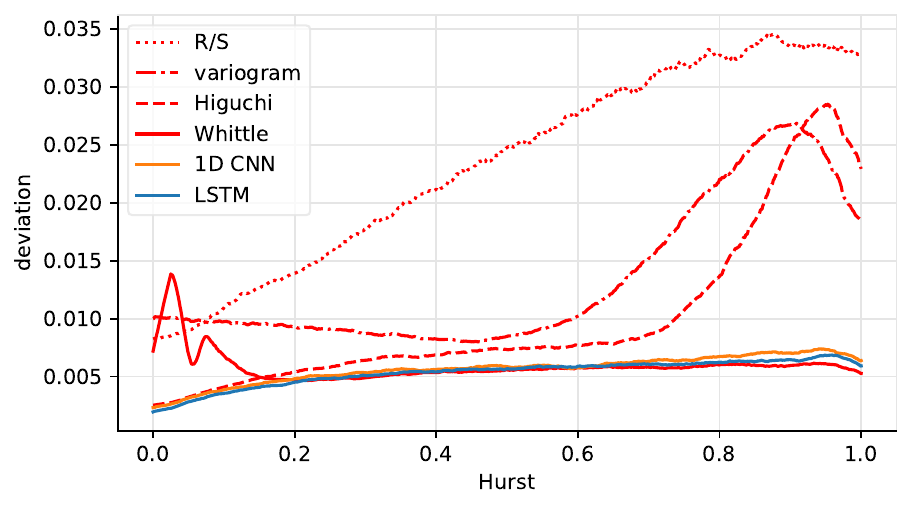}
        \subcaption{Standard deviation $\sigma_{0.025}$.}
     \end{subfigure}
     \hspace*{6mm}
     \vskip 3mm
     \caption{Empirical bias and deviation of the different fBm estimators by Hurst value, measured on sequences of length 12800. The neural models were fine tuned until $n=12800$. The bias of the R/S estimator ranges from 0.125 to -0.075, and was truncated on the plot.}
    \label{fig:fBm-bias_dev}
\end{figure*}
    \begin{figure*}[!ht]
     \vskip 0.9mm
     \centering
     \begin{subfigure}[b]{.465\textwidth}
        \centering
        \includegraphics[width=\textwidth]{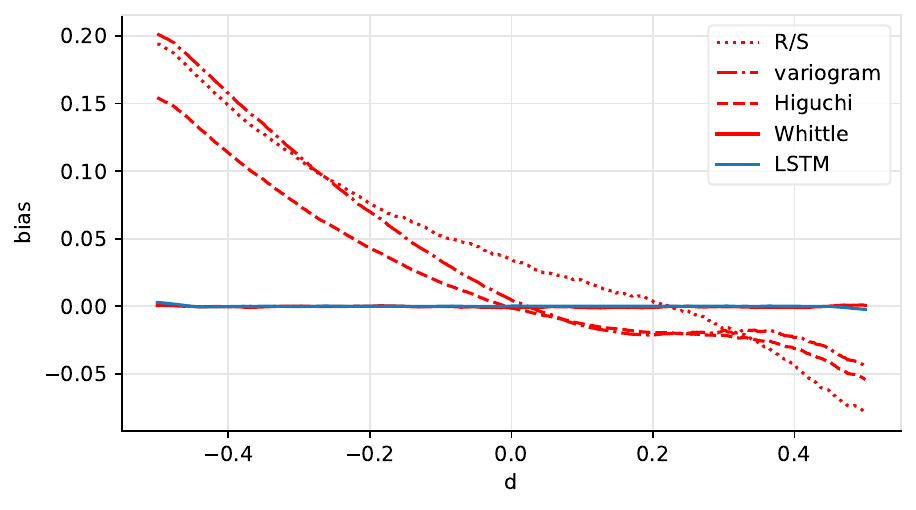}
        \subcaption{Empirical bias $b_{0.025}$.}
        \label{fig:ARFIMA-bias}
     \end{subfigure}
     \hfill
     \begin{subfigure}[b]{.465\textwidth}
        \centering
        \includegraphics[width=\textwidth]{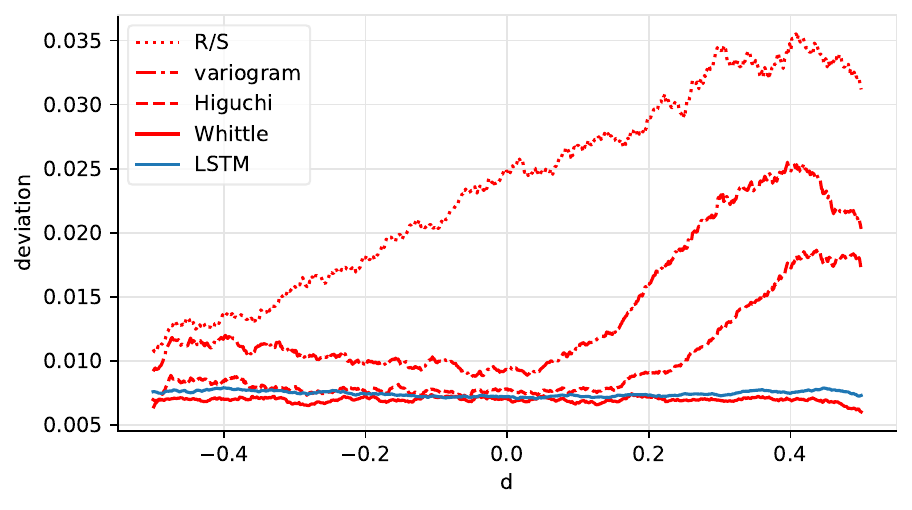}
        \subcaption{Standard deviation $\sigma_{0.025}$.}
        \label{fig:ARFIMA-dev}
     \end{subfigure}
     \hspace*{6mm}
     \vskip 3mm
     \caption{Empirical bias and deviation of the $\text{ARFIMA}(0,d,0)$ estimators by $d$. Measured on sequences of length 12800.}
    \label{fig:ARFIMA-bias_dev}
\end{figure*}
    \begin{figure*}[!ht]
    \vskip 0.9mm
    \centering
    \hspace*{0.75mm}
    \begin{subfigure}[b]{.465\textwidth}
        \centering
        \includegraphics[width=0.97\textwidth]{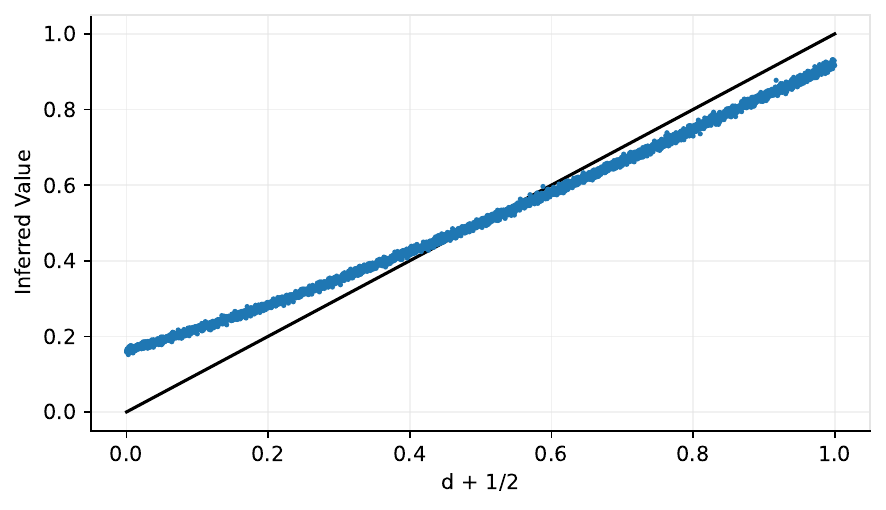}
        \subcaption{Trained on fBm, inferring on $\text{ARFIMA}$ sequences.}
    \end{subfigure}
    \hfill
    \begin{subfigure}[b]{.465\textwidth}
        \centering
        \includegraphics[width=0.99\textwidth]{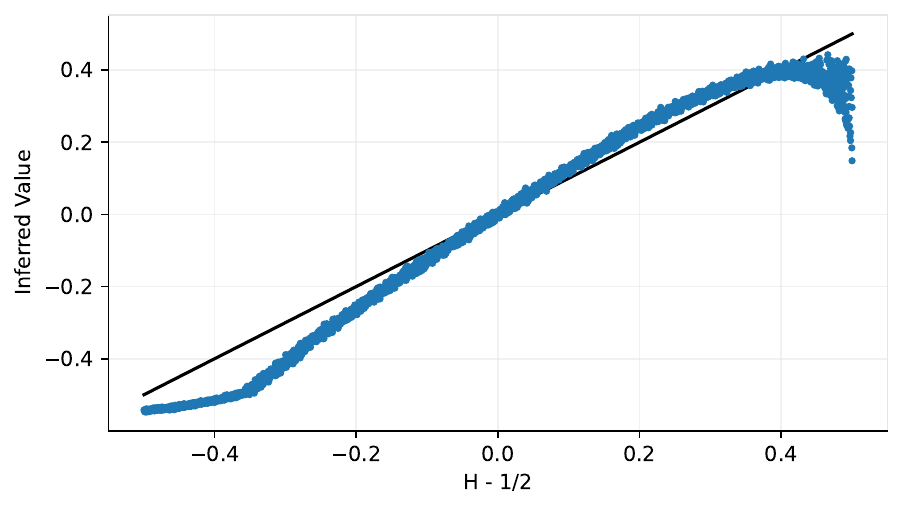}
        \subcaption{Trained on $\text{ARFIMA}$, inferring on fBm sequences.}
    \end{subfigure}
    \hspace*{-1.2mm}
    \hspace*{6mm}
    \vskip 3mm
    \caption{Scatterplots of $M_{\text{LSTM}}$ model inferences on cross-processes of length 12800. The fBm model was fine-tuned on sequences up to length 12800, and the $\text{ARFIMA}(0,d,0)$ model was trained on sequences of length 12800.}
    \label{fig:ARFIMA_fBm_cross_test}
\end{figure*}  

  \subsection{Evaluating neural fBm Hurst estimators}\label{sec:fBm_results}
    Due to self-similarity properties of fractional Brownian motion, and the stationarity of increments, a standardizing layer in the network architecture enables us to train only on realizations $\Delta FBM(H, n, 0, n)$.
    This simplified procedure yields an estimator that, in case of inference, is universally efficient regardless of the equidistant time-scale we choose, and regardless of the terminal time of the targeted process.
    Consequently, we generated sequences for training and evaluation from $\Delta \FBM(H, n, 0, n)$, where $H \sim U(0, 1)$ is random, and included the standardization layer in our models.

    In the first experiment we fine-tuned the neural models $M_{\text{conv}}$ and $M_{\text{LSTM}}$ up to $n=12800$.
    An initial training phase on trajectories of length $n=100$ was performed on $200$ and $100$ virtual epochs each containing $\num{100000}$ sequences for $M_{\text{conv}}$ and $M_{\text{LSTM}}$ respectively.
    The models were fine-tuned on $n=$ $200$, $400$, $800$, $\num{1600}$, $\num{3200}$, $\num{6400}$ and $12800$ for an additional $20$ and $10$ virtual epochs for $M_{\text{conv}}$ and $M_{\text{LSTM}}$ respectively.
    At the end we got a model which was trained on all of the sequence lengths in the list, albeit not all at the same time.
    The results can be found in Table \ref{table:fBm-MSE} and Figure \ref{fig:fBm-MSE_a}, where the loss for sequence length of $n$ was measured after fine tuning on trajectories of length $n$.
    The neural models outperform all baseline estimators especially as the sequence length increases.
    It is worth noting that while $M_{\text{LSTM}}$ achieved slightly smaller loss than $M_{\text{conv}}$, 
    it requires somewhat 
    more computational resources.  
    Figure \ref{fig:fBm-bias_dev} compares the empirical bias and standard deviation of the final fine-tuned neural models and the baseline estimators as a function on the Hurst parameter.
    \begin{table}[H]
        \caption{MSE losses of different fBm Hurst-estimators by sequence length. To enable direct comparisons with other solutions in the literature, we also included the performance of $M_{\text{LSTM}}$ where only shift invariance is ensured by turning off the standardizing layer ($M_{\text{LSTM}}^*$), here the training and evaluation was performed on $\Delta \FBM(H, n, 0, 1)$.}
        \centering
        \resizebox{0.48\textwidth}{!}{
        \begin{tabular}{lccccccc}
    & \multicolumn{7}{c}{$MSE$ loss ($\times 10^{-3}$)}  \\
    \cmidrule(r{4pt}){2-8}
    $\stackrel{\text{seq.}}{\text{len.}}$ & R/S & variogram & Higuchi & Whittle & $M_{\text{conv}}$ & $M_{\text{LSTM}}$ & $M_{\text{LSTM}}^*$\\ 
    \toprule
    $100$ & $27.6$ & $9.30$ & $10.6$ & $4.33$ & $4.27$ & $\textbf{4.07}$ & $\textit{0.214}$ \\ 
    $200$ & $18.9$ & $5.05$ & $4.21$ & $2.00$ & $1.99$ & $\textbf{1.91}$ & $\textit{0.0826}$\\ 
    $400$ & $13.9$ & $2.92$ & $1.99$ & $1.00$ & $0.959$ & $\textbf{0.917}$ & $\textit{0.0366}$\\
    $800$ & $10.8$ & $1.75$ & $1.05$ & $0.540$ & $0.476$ & $\textbf{0.453}$ & $\textit{0.0141}$\\
    $1600$ & $8.62$ & $1.09$ & $0.593$ & $0.324$ & $0.240$ & $\textbf{0.224}$ & $\textit{0.0072}$\\
    $3200$ & $6.74$ & $0.724$ & $0.360$ & $0.225$ & $0.122$ & $\textbf{0.114}$ & $\textit{0.0037}$\\
    $6400$ & $5.57$ & $0.502$ & $0.229$ & $0.179$ & $0.063$  & $\textbf{0.058}$ & $\textit{0.0029}$\\ 
    $12800$ & $4.70$ & $0.365$ & $0.155$ & $0.157$ &  $0.033$  & $\textbf{0.030}$ & $\textit{0.0032}$\\
    \bottomrule
\end{tabular}}
        \label{table:fBm-MSE}
    \end{table}
    
    We evaluated the empirical consistency of $M_{\text{LSTM}}$ trained exclusively on certain length sequences.
    We can see the results in Table \ref{table:consistency}, and Figure \ref{fig:fBm-MSE_b}.
    Trained on shorter sequences the performance of $M_{\text{LSTM}}$ improved when tested on longer sequences, but not as fast as $M_{\text{LSTM}}$ variants trained on longer sequences.
    $M_{\text{LSTM}}$ variants trained on longer sequences still performed well on shorter sequences, but not as well as dedicated variants.

    \begin{table}[H]
        \caption{MSE losses of LSTM-based fBm Hurst-estimators trained on different sequence lengths.}
        \centering
        \begin{tabular}{lccccccc} 
    {\centering \raisebox{-2ex}{train}} & \multicolumn{7}{c}{$MSE$ loss by validation seq. len. ($\times 10^{-3}$)}  \\
    \cmidrule(r{4pt}){2-8}
     {\centering $\stackrel{\text{seq.}}{\text{len.}}$} & $100$ & $200$ & $400$ & $800$ & $1600$ & $3200$ & $6400$\\ 
    \toprule
    $100$ & $4.14$ & $2.17$ & $1.41$ & $1.09$ & $0.962$ & $0.918$ & $0.915$\\
    $200$ & $4.21$ & $1.88$ & $0.947$ & $0.528$ & $0.344$ & $0.264$ & $0.231$\\
    $400$ & $4.78$ & $2.02$ & $0.940$ & $0.477$ & $0.281$ & $0.196$ & $0.161$\\
    $800$ & $4.80$ & $2.00$ & $0.913$ & $0.443$ & $0.230$ & $0.134$ & $0.0888$\\
    $1600$ & $5.01$ & $2.11$ & $0.952$ & $0.447$ & $0.220$ & $0.113$ & $0.0617$\\
    $3200$ & $5.44$ & $2.23$ & $0.972$ & $0.454$ & $0.221$ & $0.111$ & $0.0608$\\
    $6400$ & $5.59$ & $2.30$ & $1.01$ & $0.471$ & $0.229$ & $0.121$ & $0.0692$\\
    \bottomrule
\end{tabular}
        \label{table:consistency}
    \end{table}

    \subsubsection{Testing on real-world data}
\begin{figure}[H]
        \centering
        \includegraphics[width=0.48\textwidth]{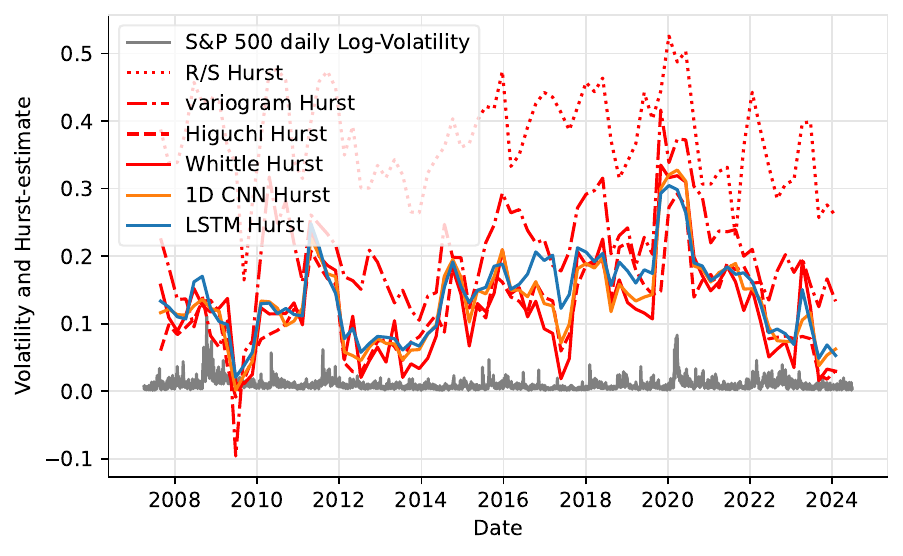}
        \caption{The figure shows Hurst-estimates for the daily S\&P 500 log-volatility, calculated from 15-minute log-returns. Estimates use 252-day (one year) sliding windows with 189-day overlaps.}
        \label{fig:volatility}
    \end{figure}

    To evaluate the real-world performance of our neural fBm Hurst-estimators, we analyzed historical data from the S\&P 500 stock market index.
    Volatility in financial markets can be modeled using fBm with Hurst exponent $H\approx 0.1$ \citep{volatility}.
    Volatility is often calculated with overlapping sliding windows, but this increases the correlation between the neighboring values and skews the Hurst-estimates to be around $0.5$.
    To avoid this, we calculated the daily volatility from 15 minute log-returns, which enabled us to estimate the Hurst parameter in yearly windows, without overlaps in the volatility calculation.
    Figure \ref{fig:volatility} shows the results
    , indicating that $M_{\text{LSTM}}$ and $M_{\text{conv}}$ effectively capture the temporal dynamics of market volatility, correlating well with traditional methods, most closely with Whittle's method, which was the best-performing baseline statistical estimator 
    in the synthetic measurements.
    The neural models yield Hurst-estimations in the $(0,\, 0.3)$ range, in the anti-persistent Hurst regime. 

    \subsubsection{Inference times}
    In our experimental setup, the neural network estimators demonstrated a favorable balance between accuracy and speed at inference.
    The advantage was only partly due to GPU usage, as suggested by the overall inference times shown in Table \ref{table:running_times}.
    The indicated inference times depend on various factors, including the implementation and available hardware resources; they reflect the specific setup used at the time of writing.
    
    \begin{table}[H]
        \caption{Overall running times for fBm Hurst-inference measured on $10000$ sequences of length $3200$.}
        \centering
        \resizebox{0.48\textwidth}{!}{
        \begin{tabular}{cccccccc}
    \toprule
    \multicolumn{6}{c}{CPU} & \multicolumn{2}{c}{GPU}  \\
    \cmidrule(r{4pt}){1-6} \cmidrule(r{4pt}){7-8}\textbf{}
    R/S & variogram & Higuchi & Whittle & $M_{\text{conv}}$ & $M_{\text{LSTM}}$ & $M_{\text{conv}}$ & $M_{\text{LSTM}}$\\ 
    \midrule
    $7\text{s}$ & $38\text{m}\,42\text{s}$ & $8\text{s}$ & $2\text{h}\,54\text{m}$ & $1\text{m}\,32\text{s}$ & $5\text{m}\,43\text{s}$ & $10\text{s}$ & $9\text{s}$\\ 
    \bottomrule
\end{tabular}

        \label{table:running_times}
        }
    \end{table}

  \subsection{Evaluating neural ARFIMA parameter estimators}\label{sec:ARFIMA_results}
    We trained $M_{\text{LSTM}}$ models for estimating the parameter $d$ of the $\text{ARFIMA}(0,d,0)$ process.
    These models contain no standardization, and work with the input sequence, not the increments.
    The models were trained on sequences of length $200$, $400$, $800$, $1600$, $3200$, $6400$ and $12800$.
    Classical Hurst estimation techniques were evaluated for the inference of $d$ as described in Section \ref{subsec:arfima}.
    By adjusting the spectrum, Whittle's method was calibrated specifically for the ARFIMA $d$-estimation.
    The results are presented in Table \ref{table:ARFIMA-MSE} and Figure \ref{fig:ARFIMA-bias_dev}.

    \begin{table}[H]
        \vskip 2.2mm
        \caption{MSE losses of different $\text{ARFIMA}(0,d,0)$ $d$-estimators by sequence length.}
        \centering
        \resizebox{0.48\textwidth}{!}{
        \begin{tabular}{lccccc}
    & \multicolumn{5}{c}{$MSE$ loss ($\times 10^{-3}$)}  \\
    \cmidrule(r{4pt}){2-6}
    seq. len. & R/S & variogram & Higuchi & Whittle  & $M_{\text{LSTM}}$ \\ 
    \toprule
    $100$   & $33.1$ & $17.3$ & $14.8$ & $9.51$   & $\textbf{5.96}$  \\ 
    $200$   & $24.0$ & $12.6$ & $8.33$ & $4.00$   & $\textbf{3.03}$  \\
    $400$   & $18.6$ & $9.83$ & $5.67$ & $1.82$   & $\textbf{1.54}$  \\
    $800$   & $14.9$ & $8.11$ & $4.67$ & $0.846$  & $\textbf{0.787}$ \\
    $1600$  & $12.6$ & $7.70$ & $4.24$ & $0.401$  & $\textbf{0.390}$ \\
    $3200$  & $9.90$ & $7.00$ & $3.80$ & $0.200$  & $\textbf{0.199}$ \\
    $6400$  & $8.64$ & $6.95$ & $3.80$ & $\textbf{0.0960}$ & $0.104$ \\
    $12800$ & $7.51$ & $6.94$ & $3.75$ & $\textbf{0.0487}$ & $0.0552$\\
    \bottomrule
\end{tabular}
        \label{table:ARFIMA-MSE}
        }
    \end{table}
    
    \subsubsection{Stress-testing with ARFIMA}
    We conducted cross-tests on $M_{\text{LSTM}}$ models, by training them on ARFIMA and fBm trajectories, and then evaluating them against the alternate process.
    The ARFIMA $d$-estimator was trained on $n=12800$, and the fBm $H$-estimator was fine-tuned up to $n=12800$.
    In both cases the models were tested on sequences of length $12800$.
    As Figure \ref{fig:ARFIMA_fBm_cross_test} shows, the models perform remarkably, with minor asymmetric bias with respect to the parameter range.
    This suggests that the models either capture the decay rate of autocovariance of fractional noise or some fractal property of sample paths.

  \subsection{Evaluating neural fOU parameter estimators}\label{subsec:fOU_results}
    Estimating the parameters of the fractional Ornstein-Uhlenbeck process is significantly more difficult.
    We evaluated ${\mathcal{M}}_{\text{LSTM}}$ on the estimation of the Hurst parameter of fOU.
    Here $M_{\text{LSTM}}$ does not work with the increments, but includes standardization to ensure scale and shift invariance.
    As we stated in Section \ref{sec:fOU_bg} these invariances enable training on $\FOU(\eta, H, \alpha, 0, 1)$ without the loss of generality.
    Additionally we evaluated the quadratic generalized variation (QGV) estimator for the Hurst parameter, as described in \citep{brouste2011}.
    We trained $M_{\text{LSTM}}$ on sequences of length $200$, $400$, $800$, $1600$, $3200$, $6400$ and $12800$
    with the fine-tuning technique similar to the one in Section \ref{sec:fBm_results}.
    We generated the sequences for training and evaluation of the Hurst estimators from $\FOU(\eta, H, \alpha, 0, 1)$, where $H \sim U(0, 1), \alpha \sim \Exp(100)$ and $ \eta \sim N(0, 1)$ are random.
    
    \begin{table}[H]
        \vskip 2.2mm
        \caption{Performance metrics of different fOU Hurst-estimators by sequence length.}
        \centering

\begin{tabular}{lcccccc}
    & \multicolumn{2}{c}{$MSE$ loss ($\times 10^{-3}$)}  & \multicolumn{2}{c}{$\hat{b}_{0.025}$ ($\times 10^{-3}$)} & \multicolumn{2}{c}{$\hat{\sigma}_{0.025}$ ($\times 10^{-2}$)} \\
    \cmidrule(r{4pt}){2-3} \cmidrule(r{4pt}){4-5} \cmidrule(r{4pt}){6-7}
    seq. len. & QGV & $M_{\text{LSTM}}$ & QGV & $M_{\text{LSTM}}$ & QGV & $M_{\text{LSTM}}$\\ 
    \midrule
    $100$   & $41.0$ & $\textbf{3.33}$  & $106$  & $\textbf{9.51}$  & $9.86$ & $\textbf{5.49}$ \\ 
    $200$   & $34.2$ & $\textbf{1.71}$  & $97.1$ & $\textbf{4.85}$  & $8.07$ & $\textbf{3.98}$ \\ 
    $400$   & $29.4$ & $\textbf{0.908}$ & $86.4$ & $\textbf{2.58}$  & $6.92$ & $\textbf{2.89}$ \\ 
    $800$   & $25.0$ & $\textbf{0.486}$ & $76.2$ & $\textbf{1.49}$  & $5.88$ & $\textbf{2.14}$ \\ 
    $1600$  & $20.6$ & $\textbf{0.261}$ & $65.1$ & $\textbf{0.814}$ & $5.09$ & $\textbf{1.56}$ \\ 
    $3200$  & $16.3$ & $\textbf{0.141}$ & $53.8$ & $\textbf{0.431}$ & $4.37$ & $\textbf{1.14}$ \\  
    $6400$  & $12.6$ & $\textbf{0.074}$ & $43.7$ & $\textbf{0.306}$  & $3.68$ & $\textbf{0.834}$ \\ 
    $12800$ & $9.28$ & $\textbf{0.042}$ & $33.6$ & $\textbf{0.191}$  & $3.06$ & $\textbf{0.616}$ \\ 
\end{tabular}
        \label{table:fOU-Hurst}
    \end{table}

\subsubsection{Stress-testing with OU}
\begin{figure}[H]
        \centering
        \includegraphics[width=0.48\textwidth]{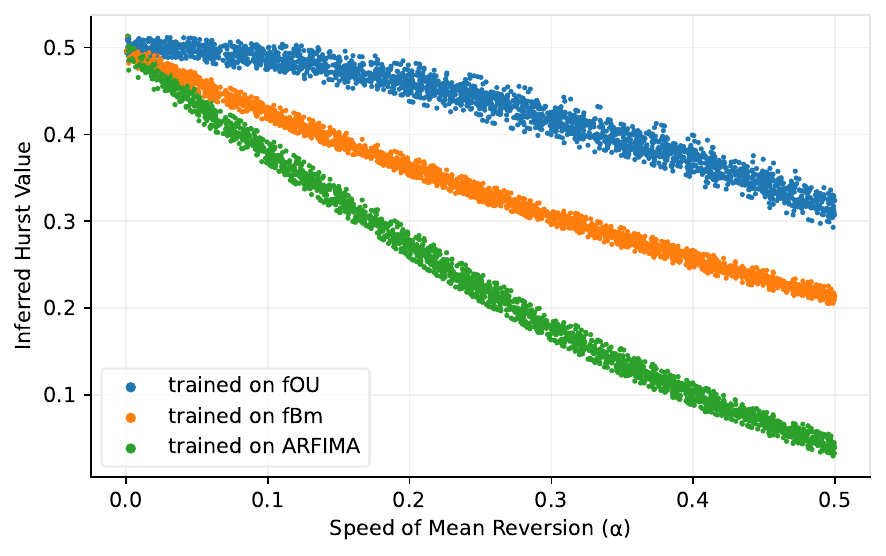}
        \caption{Scatterplot of Hurst estimations by $M_{\text{LSTM}}$ models fine-tuned up to 12800 length sequences, inferring on Ornstein–Uhlenbeck processes of length 12800.}
        \label{fig:OU_scatter}
    \end{figure}

    We stress-tested the $M_{\text{LSTM}}$ Hurst-estimators trained on fBm, fOU and ARFIMA processes (in the case of ARFIMA, originally trained for estimating $d$) on the standard Ornstein-Uhlenbeck process $\FOU(0, 0.5, \alpha, 0, 1)$; results are shown in Figure \ref{fig:OU_scatter}.
    The inferred value at $\alpha=0$ is $0.5$ as expected since the model receives an input that it already encountered in the learning phase.
    As the parameter $\alpha$ increases, all models return values tending towards zero.
    This could suggest that larger $\alpha$ values have a similar effect to smaller Hurst values, and sequences generated with a larger speed of mean reversion resemble trajectories that have greater anti-persistence.
    Moreover when $\alpha > 0$, the Ornstein-Uhlenbeck autocovariance shows exponential decay, contrary to the power decay associated with the data that the models were calibrated on.
    However, the fOU Hurst-estimator does stay around $0.5$ longer, corresponding to the $\alpha$ range it encountered during training.

\section{Conclusion}
    In this work, we have demonstrated the effectiveness of sequence-processing neural networks as powerful tools for estimating statistical parameters associated with long memory, such as the Hurst exponent. 
    We have demonstrated the superior performance and consistency of purely neural network-based models over several commonly used statistical techniques in the context of fBm, fOU, and ARFIMA processes. 
    This suggests that complex statistical descriptors can be effectively represented by relatively simple neural networks, a phenomenon that deserves further investigation.
    
    The compact size of the neural networks used in our study enables fast inference, and the quality of the estimates they provide is on par with, or superior to, the most precise yet resource-intensive existing methods.
    A limitation of our arguments is the length of the sequences used in the measurements, nevertheless, in practice, the sequence lengths dealt with are sufficient in the vast majority of cases.
    
    We believe that the proposed parameter estimators of stochastic processes, based on recurrent neural networks, with their usability and flexibility, form a good basis for the estimation methods that can be used for more intricate processes.





\begin{ack}
The research was supported by the Hungarian National Research, Development and Innovation Office within the framework of the Thematic Excellence Program 2021 -- National Research Sub programme: ``Artificial intelligence, large networks, data security: mathematical foundation and applications'' and the Artificial Intelligence National Laboratory Program (MILAB), and the Hungarian National Excellence Grant 2018-1.2.1-NKP-00008.
We would also like to thank GitHub and neptune.ai for providing us academic access.
\end{ack}


\bibliography{param_estim_ECAI2024}

\newpage
\appendix



\part*{Appendix} 
    \begin{itemize}[align=left]
        \item[\textbf{Section \ref{sec_scaling}:}] Highlights the significance of scale-invariance within our neural models.
        \item[\textbf{Section \ref{sec_more_stress}:}] Explores how fBm Hurst-estimators perform when applied to processes other than what they were trained on.
        \item[\textbf{Section \ref{sec_proc_prop}:}] Provides potential insights into which process properties the neural Hurst-estimators are sensitive to, including memory, self-similarity, and the fractal dimension of the trajectories.
        \item[\textbf{Section \ref{sec_tables}:}] Contains tables with additional numerical results from our measurements.
        \item[\textbf{Section \ref{sec_stat_tests}:}] Elaborates on the process generators utilized for model training, and evaluates our neural estimators using sequences generated independently of our project.
        \item[\textbf{Section \ref{realizations}:}] Presents figures of various process realizations and a typical training curve.        
    \end{itemize}

\section{Validating model scale-invariance} \label{sec_scaling}
\begin{figure}[H]
        \centering
        \vspace*{-2.5mm}
        \includegraphics[width=0.48\textwidth]{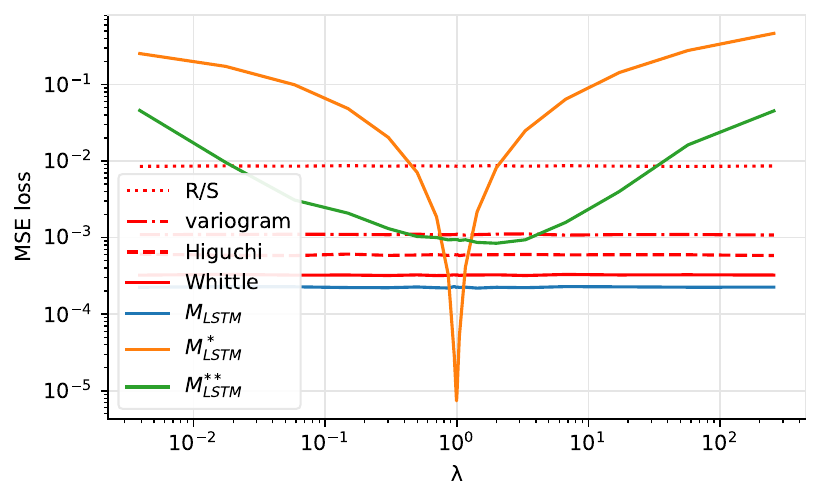}
        \caption{MSE losses of different fBm Hurst-estimators on realizations  $\lambda\cdot\Delta\FBM(H,1600,0,1)$ as a function of $\lambda$. $M_{\text{LSTM}}^*$ and $M_{\text{LSTM}}^{**}$ contain no standardizing layer and were trained on \mbox{sequences} $\Delta \FBM(H, 1600, 0, 1)$ and $\lambda\cdot \Delta \FBM(H, 1600, 0, 1)$: $\lambda \sim \Exp(1)$ respectively. $M_{\text{LSTM}}$ was also trained on realizations of length $1600$.}
        \label{fig:scaling}
    \end{figure}

    Scale-invariance is a central feature of our approach.
    To underscore its significance in developing reliable neural Hurst-estimators, we designed an experiment to assess various estimators across fBm realizations sampled from a range of time-horizons.
    Considering that adjusting the sample time-horizon is equivalent to scaling by a factor of $\lambda$, we calculated the MSE losses of the estimators for fBm realizations scaled by various $\lambda$ values (Figure \ref{fig:scaling}).
    Our $M_{\text{LSTM}}$ model incorporates a standardizing layer and, as anticipated, demonstrates scale-independent performance.
    Additionally, none of the baseline statistical estimators indicate any scale-dependency.
    In Section \ref{sec:fBm_results} of the main article we include the performance of $M_{\text{LSTM}}^*$ in Table \ref{table:fBm-MSE}. 
    $M_{\text{LSTM}}^*$ is trained and evaluated on $\Delta \FBM(H, 1600, 0, 1)$ realizations, and outperforms $M_{\text{LSTM}}$ locally by not including the standardizing layer.
    Figure \ref{fig:scaling} indicates the limited range of applicability of $M_{\text{LSTM}}^*$ if the time horizon is also an unknown parameter of the process, which is typically the case in practice.
    In order to further highlight the importance of the standardizing layer, we trained a model $M_{\text{LSTM}}^{**}$ on $\lambda\cdot \Delta \FBM(H, 1600, 0, 1)$ realizations where we sampled $\lambda$ from $\Exp(1)$.
    While the performance of $M_{\text{LSTM}}^{**}$ is less extremely localized around $\lambda\approx 1$, as opposed to $M_{\text{LSTM}}$, it does not outperform the baseline estimators at any scale.

\section{Stress-testing Hurst-estimators} \label{sec_more_stress}
  \subsection{The impact of noise and smoothing on fBm}
    In this section, we examine the effects of artificially increasing or decreasing the roughness of fBm realizations before performing inference.
    Our aim is to gain insights into how different Hurst-estimators respond to modifications in data roughness.
    By increasing the roughness through the addition of white noise, we can understand the resilience of inference algorithms under noisy conditions, mirroring real-world data imperfections.
    Conversely, decreasing roughness via a moving average filter helps evaluate the sensitivity of our models to smoothed data, often encountered due to aggregation or preprocessing steps.

\begin{figure*}[!ht]
     \centering
     \begin{subfigure}[b]{.465\textwidth}
        \centering
        \includegraphics[width=\textwidth]{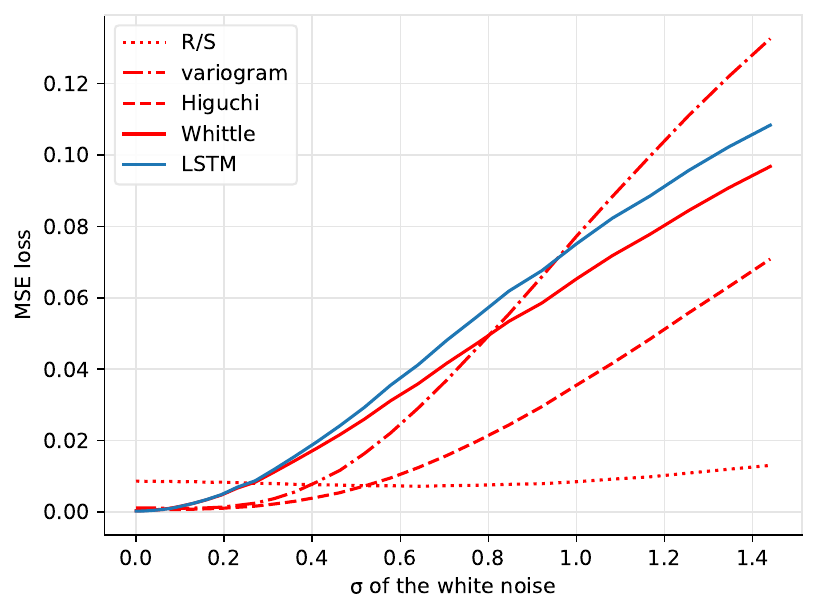}
        \subcaption{MSE losses by the deviation of the white noise.}
        \label{fig:white_noise_plot}
     \end{subfigure}
     \hfill
     \begin{subfigure}[b]{.46\textwidth}
        \centering
        \includegraphics[width=\textwidth]{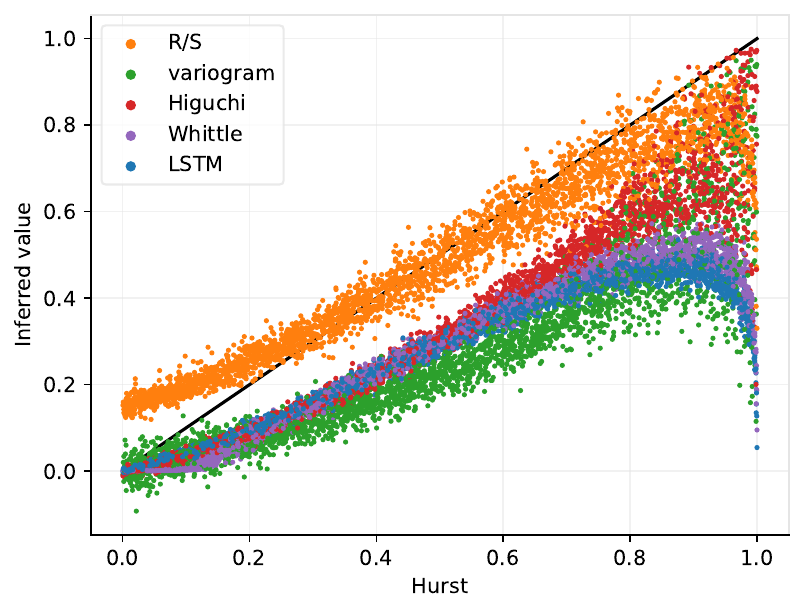}
        \subcaption{Scatterplot at $\sigma=1$.}
        \label{fig:white_noise_scatter}
     \end{subfigure}
     \hspace*{6mm}
     \vskip 3mm
     \caption{The effect of adding Gaussian white noise to the fBm realizations before inference. }
    \label{fig:white_noise}
\end{figure*}

    Figure \ref{fig:white_noise} illustrates the effects of adding Gaussian white noise to fBm realizations before inference.
    Note that this scenario can be considered a special case of the stress-test in Subsection \ref{subsec::memoryvsbox}, given that as $H \to 0$, fBm increments resemble white noise.
    Figure \ref{fig:white_noise_plot} shows how the error of various estimators changes with increasing noise strength ($\sigma$).
    $M_{\text{LSTM}}$ reacts to increasing noise levels quite rapidly, closely following Whittle's method. 
    It is noteworthy that the $R/S$ statistic is mostly agnostic to the amount of the added white noise.

\begin{figure*}[!ht]
     \vskip 0.9mm
     \centering
     \begin{subfigure}[b]{.465\textwidth}
        \centering
        \includegraphics[width=\textwidth]{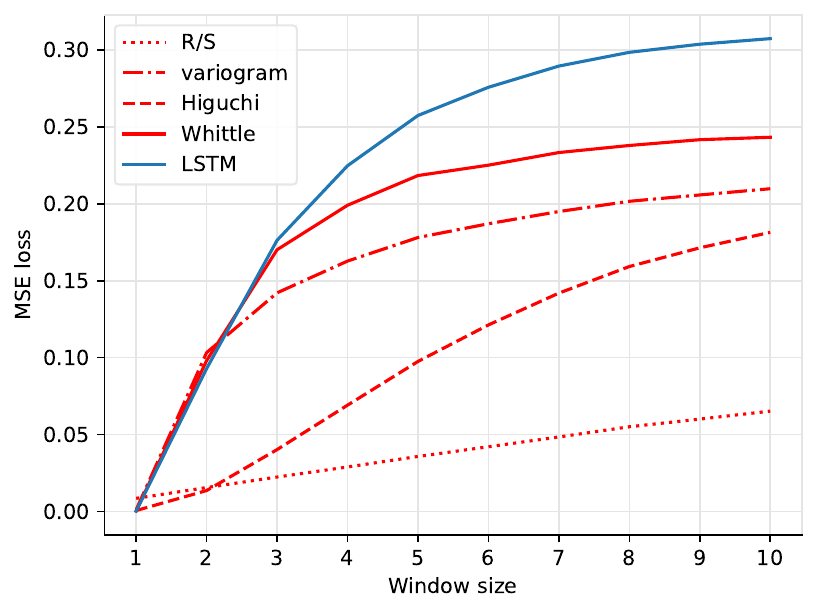}
        \subcaption{MSE losses by the smoothing window size.}
        \label{fig:smoothing_plot}
     \end{subfigure}
     \hfill
     \begin{subfigure}[b]{.46\textwidth}
        \centering
        \includegraphics[width=\textwidth]{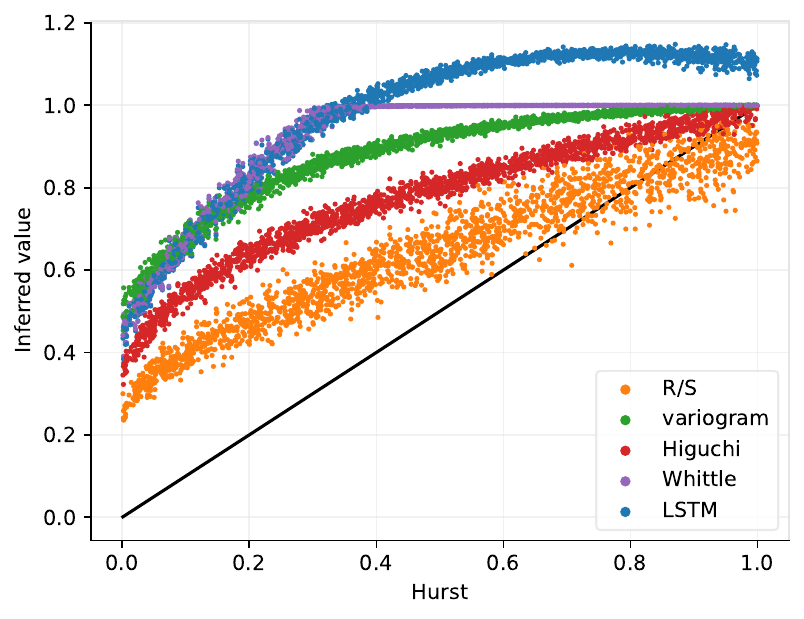}
        \subcaption{Scatterplot at window size $5$.}
        \label{fig:smoothing_scatter}
     \end{subfigure}
     \hspace*{6mm}
     \vskip 3mm
     \caption{The effect of smoothing $\FBM(H, 1600, 0, 1600)$ realizations before inference using a moving average window at stride 1.}
     \vskip 3mm
    \label{fig:smoothing}
\end{figure*}

    Figure \ref{fig:smoothing} shows how estimation errors change with increasing moving average window sizes, indicating that when the smoothing windows overlap, there is a notable increase in the estimated Hurst parameter.
    This suggests that the estimators perceive smoothed data as exhibiting stronger long-range dependence.
    Among the estimators tested, $M_{\text{LSTM}}$ responds most rapidly.
    In contrast, the $R/S$ statistic reacts more slowly but is not entirely insensitive to the size of the smoothing window.

\begin{figure*}[!ht]
     \vskip 0.9mm
     \centering
     \begin{subfigure}[b]{.465\textwidth}
        \centering
        \includegraphics[width=\textwidth]{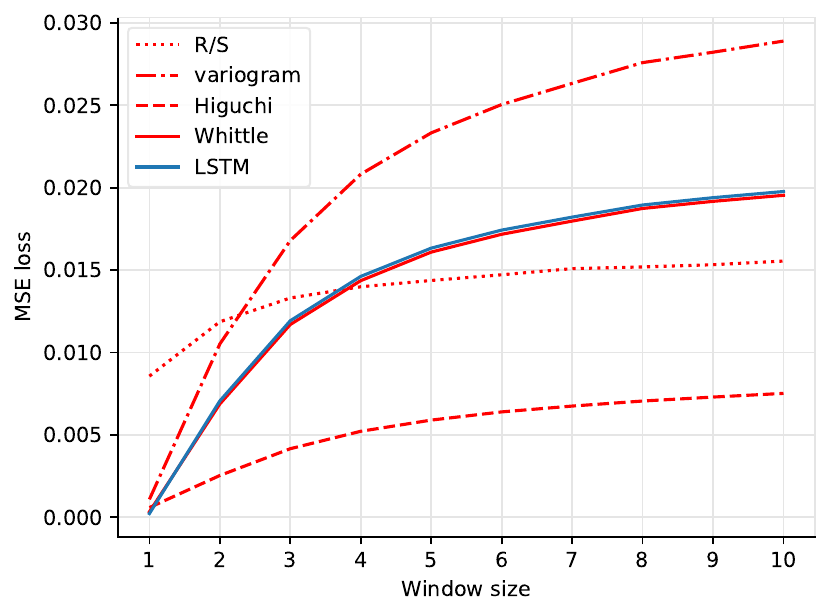}
        \subcaption{MSE losses by the smoothing window size.}
        \label{fig:smoothing_plot_no_overlap}
     \end{subfigure}
     \hfill
     \begin{subfigure}[b]{.45\textwidth}
        \centering
        \includegraphics[width=\textwidth]{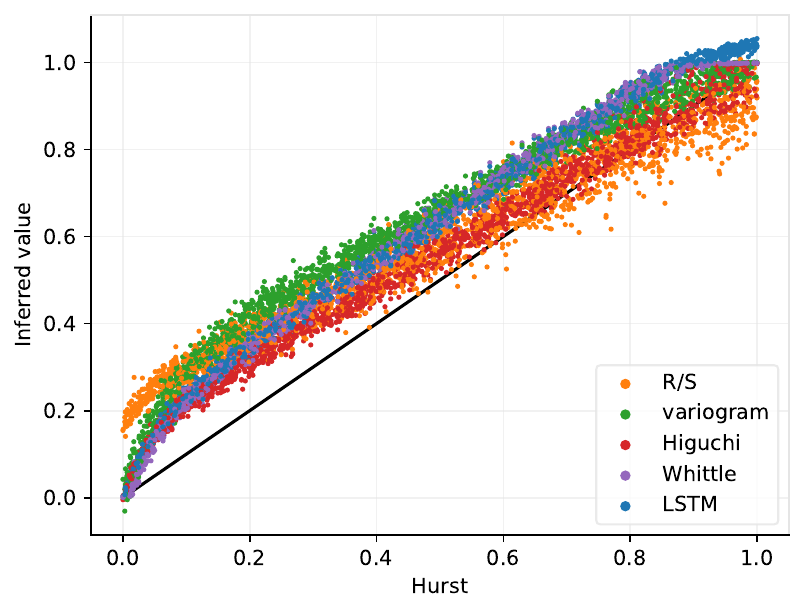}
        \subcaption{Scatterplot at window size $5$.}
        \label{fig:smoothing_scatter_no_overlap}
     \end{subfigure}
     \hspace*{6mm}
     \vskip 3mm
     \caption{The effect of smoothing $\FBM(H, 1600 \cdot m, 0, 1600 \cdot m)$ realizations before inference using non-overlapping moving average windows of size $m$.}
     \vskip 3mm
    \label{fig:smoothing_no_overlap}
\end{figure*}

\begin{figure}[H]
        \centering
        \includegraphics[width=0.472\textwidth]{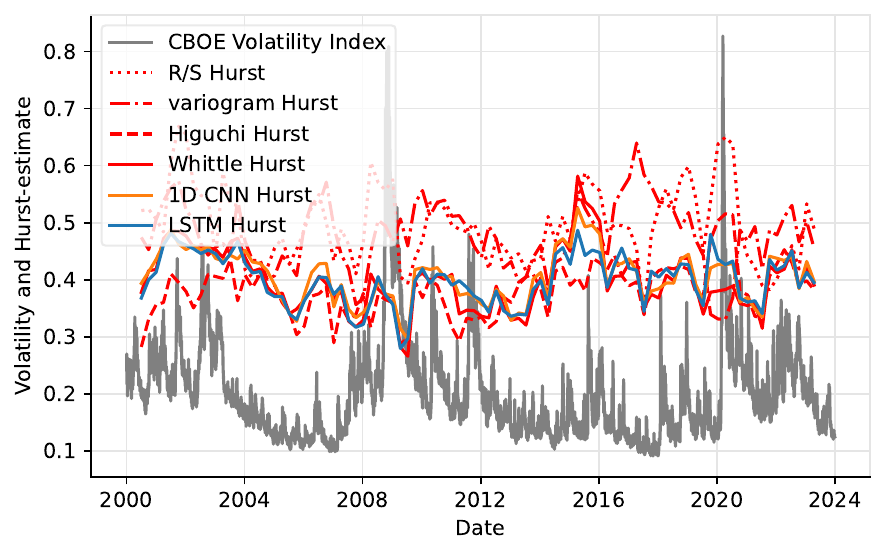}
        \caption{The figure shows the Hurst-estimate for the daily closing CBOE Volatility Index (VIX) from 2000 until 2024. Estimates use 252-day sliding windows (one year) with 126-day overlaps.}
        \label{fig:volatility_VIX}
    \end{figure}

    The overlapping of smoothing windows contributes significantly to the increase in the estimated Hurst parameter.
    As shown in Figure \ref{fig:smoothing_no_overlap}, when we use non-overlapping windows, the models still overestimate $H$, but to a much smaller degree.
    This can likely be explained by considering that overlapping windows increase autocorrelation, which directly affects the Hurst estimation.
    A similar effect can be observed when we apply our Hurst-estimators to market volatility data.
    For example, the Chicago Board Options Exchange Volatility Index (VIX), also known as the ``fear index'', is calculated from the S\&P 500 and measures the market's expectation of volatility over the next 30 days.
    Figure \ref{fig:volatility_VIX} shows what happens, when we apply the Hurst-estimators directly to this data.
    The Hurst estimations are much larger than those in Figure \ref{fig:volatility} of the main article, which also contains Hurst estimations of the volatility calculated from the S\&P 500, but within non-overlapping time windows.


  \subsection{Inference on autoregressive processes}
\begin{figure}[H]
        \centering
        \includegraphics[width=0.48\textwidth]{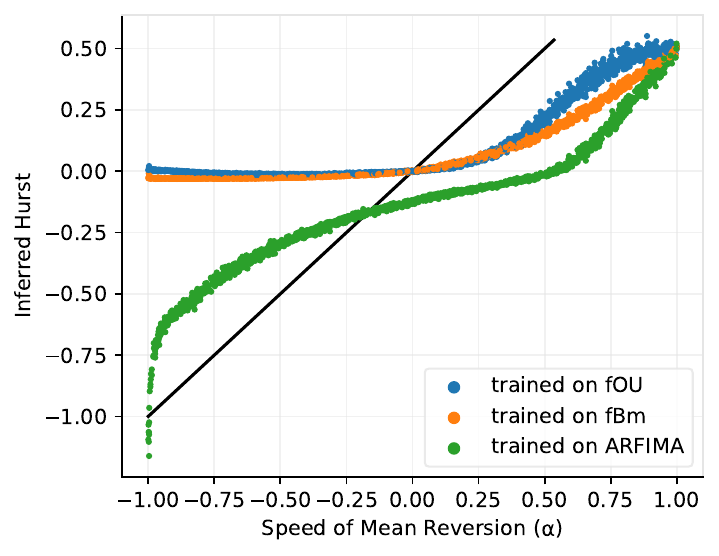}
        \caption{Scatterplot of Hurst-estimations by $M_{\text{LSTM}}$ models fine-tuned up to 12800 length sequences, inferring on autoregressive processes of order 1 and length 3200.}
        \label{fig:autoregression_scatter}
        \vspace*{0.5cm}
    \end{figure}

    We stress-tested the different $M_{\text{LSTM}}$ Hurst-estimators, by applying them to autoregressive processes of order 1, defined as:
    $$
    X_1 = w_1, \quad X_t = \alpha X_{t-1} + w_t,\, \text{for } t = 2, 3, \ldots, n,
    $$
    where $\alpha$ is the speed of mean reversion and $w_t \sim N(0, 1)$.
    The estimators were trained on fBm, fOU and ARFIMA processes (in the case of ARFIMA, originally trained for estimating parameter $d$ instead of $H$).
    Figure~\ref{fig:autoregression_scatter} presents the results of the test.

    When $\alpha$ is close to $1$, the autoregressive process approximates a standard Brownian motion, corresponding to a fractional Brownian motion with a Hurst parameter of $H = 0.5$.
    This explains the inferred values near $\alpha=1$. 
    Notably, the exponential decay of memory in autoregressive processes is fundamentally different from the long-range dependence exhibited by fractional processes. 
    As $\alpha$ deviates from $1$, the mean-reverting process moves away from a behavior resembling a random walk, and inferred values follow the change: possibly indicating that the decay of memory is not according to a power-law.
    
    The $M_{\text{LSTM}}$ models trained for Hurst-estimation on fBm and fOU processes behave similarly: as $\alpha \to 0$, the inferred Hurst values approach $0$, which is consistent with the process being dominated by white noise, analogous to fBm and fOU when $H \to 0$.
    Moreover, these estimators show persistence, continuing to produce values near 0, even when $\alpha$ becomes negative.

    In contrast, the ARFIMA-based Hurst-estimator displays a different pattern.
    While it also infers values around $H = 0.5$ for $\alpha$ values close to $1$, it does not generally yield estimates bounded from below by 0.
    This is noteworthy, as the estimator was trained on a parameter range of $d\in(-0.5, 0.5)$ corresponding to $H = d+0.5 \in (0, 1)$, meaning that the values it returns for $\alpha<0$ were outside its training range.
    A potential explanation for the discrepancy between the estimators is that ARFIMA is a more flexible model, capable of capturing both long-range dependence and heavy-tailed behavior when operating in the domain of short memory processes.
 
\section{Testing model sensitivity to key fBm properties}
\label{sec_proc_prop}
    Consider a black-box estimator such as $M_{\text{LSTM}}$, which accepts a sample path from a fractional Brownian motion characterized by an unknown parameter $H \in (0,1)$.
    The task of the estimator is to reproduce the quantity $H$.
    If the estimation is successful, one can investigate the unknown methodology that the estimator utilizes to produce its output.
    The estimator potentially captures at least three distinct characteristic quantities: the box dimension of the paths, the properties associated with memory, or the exponent of self-similarity.

  \subsection{Inference on the sum of two fBMs}\label{subsec::memoryvsbox}
    To potentially distinguish between the fractal characteristics of paths (box dimension) and the decay of auto-covariance (memory),
    let us consider two real numbers $H_1<H_2$ in the set $(0,0.5)\cup (0.5,1)$, and let $B_t^{H_1}$ and $B_t^{H_2}$ be two independent fractional Brownian motions with parameters $H_1$ and $H_2$ respectively.
    Consider the process defined by $X_t = B_t^{H_1} + B_t^{H_2}$.
    On one hand, if a Hurst-estimator captures the asymptotic decay of the autocovariance of the given input, then it should infer a value close to $H_2$.
    On the other hand, if it captures the box dimension, then it will return a value near $H_1$.
    Thus, the heuristic reasoning outlined above suggests a potential method,
    with which one can test the otherwise unknown behavior of the estimator $M_{\text{LSTM}}$, regarding its hidden estimation procedure.

    \begin{figure}[H]
        \centering
        \includegraphics[width=0.48\textwidth]{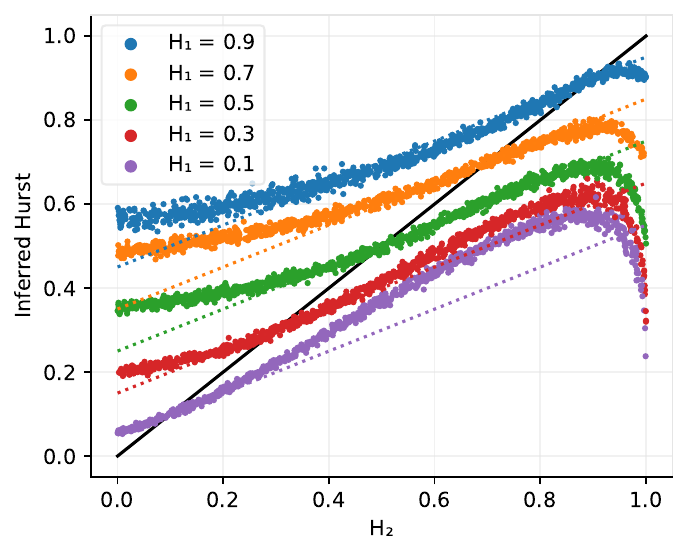}
        \caption{Scatterplot of the $H$ values estimated by $M_{\text{LSTM}}$ on fBm sum inputs with fixed $H_1$ and uniform random $H_2$ values. Series length $n=6400$. Dotted lines indicate the average of $H_1$ and $H_2$.}
        \label{hurst2_scatter}
    \end{figure}

    Driven by these motivations, after training $M_{\text{LSTM}}$ for the parameter estimation of the fBm, we tested it on the above fBm sums.
    We considered cases where $H_1$ was fixed and $H_2 \sim U(0, 1)$ was random.
    The resulting scatter plot is shown in Figure \ref{hurst2_scatter}.
    Apparently, $M_{\text{LSTM}}$ tends to infer values $\hat{H} \in (H_1, H_2)$.
    Consequently, the model appears not to learn either the box dimension or the memory explicitly but rather an intermediate quantity.
    Interestingly, when $H_2$ is near 1, the estimates provided by $M_{\text{LSTM}}$ are smaller than the average of $H_1$ and $H_2$.
    Conversely, when $H_2>0.5$ is not close to 1 and $H_1<0.5$, the estimates are larger than the average, suggesting that the memory aspect associated with $H_2$ has a greater influence.
    This behavior suggests that the estimator captures different characteristics of the process, and that in different $H_2$ ranges, different characteristics dominate the sum.
    
  \subsection{Inference on Lévy processes}\label{subsec::selfs}
    Symmetric $\alpha$-stable Lévy processes, for $\alpha \in (0,2]$, are known to exhibit self-similarity with a scaling exponent of $1/\alpha$.
    Additionally, \citet{West1982} demonstrated that for $\alpha > 0.5$, the box dimension of these processes can be expressed as $2 - 1/\alpha$.
    Therefore, when a Hurst-estimator is designed to capture properties such as box dimension or self-similarity, it should yield the value $1/\alpha$ in both contexts.
    Despite their self-similar nature, Lévy processes lack traditional memory, as their increments are independent and uncorrelated over time.
    Consequently, when an estimator is used to infer characteristics tied to memory, such as the decay of autocovariance, the resulting estimates should asymptotically approach zero, reflecting the independence of increments in these processes.

    \begin{figure}[H]
        \vspace*{-0.1cm}
        \centering
        \includegraphics[width=0.48\textwidth]{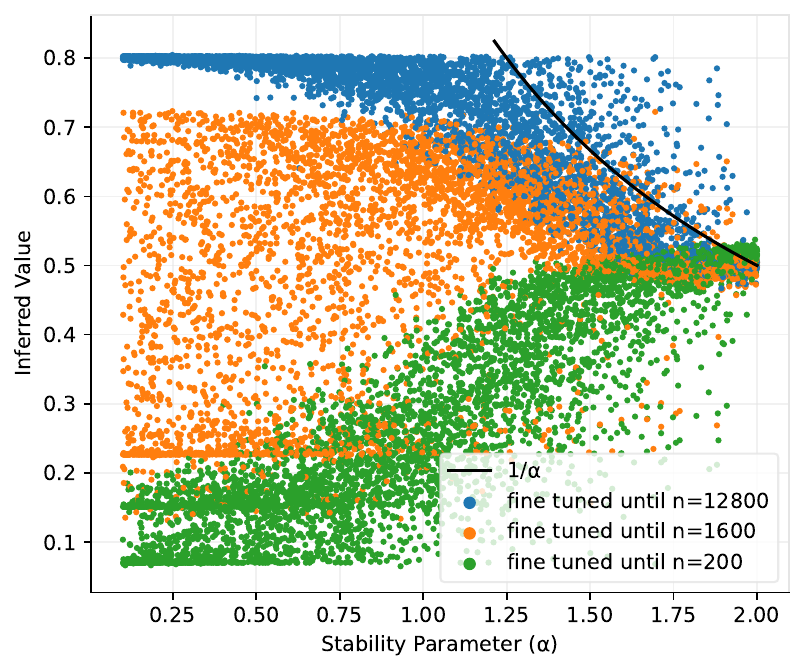}
        \caption{Scatterplot of $M_{\text{LSTM}}$ Hurst-estimators fine-tuned up to different lengths of fBm sequences inferring on Lévy processes of length 6400 and stability parameter $\alpha$.}
        \label{fig:levy_scatter}
    \end{figure}
    
    Several $M_{\text{LSTM}}$ Hurst-estimators, fine-tuned on fBm sequences of different lengths, were evaluated on $\alpha$-stable Lévy processes, with outcomes illustrated in Figure \ref{fig:levy_scatter}. 
    Notably, for $\alpha = 2$, the characteristics of an $\alpha$-stable Lévy process overlap with those of standard Brownian motion, which corresponds to a fBm with a Hurst parameter of $H=0.5$.
    Thus, the inferred value is around $0.5$ at $\alpha =2$, as anticipated.
    
    However, when moving away from $\alpha=2$, 
    The models trained on different length fBm realizations demonstrate significantly different behaviors. 
    As $\alpha$ approaches $0$, the model trained and fine-tuned on fBm sequences only up to length $n = 200$, yields estimates approaching $0$, albeit cutting off slightly above 0.
    The model fine-tuned up to $n = 1600$ displays the highest variance, yielding estimates between the other two models.
    Finally, the model fine-tuned up to $n = 12800$ gives increasing estimates as $\alpha$ decreases, somewhat resembling the $1/\alpha$ value we could expect if it captured the box dimension or self-similarity.
    Interestingly this model also cuts off at a seemingly arbitrary value near $0.8$.

    It might be possible that models trained on shorter sequences primarily learn to infer the Hurst parameter based on the decay of autocovariance.
    In contrast, as the fine-tuning progressed on longer sequences, the models may have gradually shifted their focus to other process attributes, such as box dimension or self-similarity.
    This potential shift in learned properties warrants a more thorough investigation in the future.
    
\section{Additional aggregated empirical bias and standard deviation model evaluations} \label{sec_tables}

    Tables \ref{table:fBm-bias} and \ref{table:fBm-dev} provide additional information about the fBm Hurst-estimators which appear in Table \ref{table:consistency} and Figure \ref{fig:fBm-bias_dev} in Section \ref{sec:fBm_results}.
    
    \begin{table}[H]
        \vspace*{0.5cm}
        \centering
        \caption{Absolute area under Hurst--bias curve of different fBm Hurst-estimators by sequence length.}
        \centering
        \resizebox{0.48\textwidth}{!}{
        \begin{tabular}{lccccccc}
    & \multicolumn{7}{c}{$\hat{b}_{0.025}$ ($\times 10^{-3}$)}  \\
    \cmidrule(r{4pt}){2-8}
    $\stackrel{\text{seq.}}{\text{len.}}$ & R/S    & variogram & Higuchi & Whittle        & $M_{\text{conv}}$ & $M_{\text{LSTM}}$ & $M_{\text{LSTM}}^*$\\
    \midrule
    $100$     & $116$  & $34.9$    & $58.1$ & $\textbf{10.7}$ & $12.2$            & $11.3$           & $\textit{0.712}$\\ 
    $200$     & $98.0$ & $26.5$    & $27.7$ & $6.84$          & $5.54$            & $\textbf{5.24}$  & $\textit{0.281}$\\ 
    $400$     & $87.1$ & $20.7$    & $14.5$ & $5.79$          & $2.63$            & $\textbf{2.58}$  & $\textit{0.148}$\\
    $800$     & $79.1$ & $17.0$    & $7.82$ & $5.03$          & $1.40$            & $\textbf{1.32}$  & $\textit{0.0630}$\\
    $1600$    & $71.4$ & $13.9$    & $4.63$ & $4.90$          & $0.765$           & $\textbf{0.656}$ & $\textit{0.0467}$\\
    $3200$    & $62.6$ & $11.8$    & $2.89$ & $4.88$          & $0.411$           & $\textbf{0.403}$ & $\textit{0.0239}$\\
    $6400$    & $58.0$ & $9.79$    & $1.74$ & $5.03$          & $0.241$           & $\textbf{0.211}$ & $\textit{0.0241}$\\ 
    $12800$   & $53.7$ & $8.34$    & $1.23$ & $5.04$          & $0.140$           & $\textbf{0.131}$ & $\textit{0.0326}$\\
\end{tabular}
        \label{table:fBm-bias}
        }
    \end{table}
    
    \begin{table}[H]
        \centering
        \caption{Absolute area under Hurst--deviation curve of different fBm Hurst-estimators by sequence length.}
        \centering
        \resizebox{0.48\textwidth}{!}{
        \begin{tabular}{lccccccc}
    & \multicolumn{7}{c}{$\hat{\sigma}_{0.025}$ ($\times 10^{-2}$)}  \\
    \cmidrule(r{4pt}){2-8}
    $\stackrel{\text{seq.}}{\text{len.}}$ & R/S & variogram & Higuchi & Whittle & $M_{\text{conv}}$ & $M_{\text{LSTM}}$ & $M_{\text{LSTM}}^*$\\ 
    \midrule
    $100$   & $9.62$ & $8.67$ & $8.25$  & $6.25$  & $6.16$  & $\textbf{6.01}$ & $\textit{1.42}$\\ 
    $200$   & $7.35$ & $6.33$ & $5.65$  & $4.30$  & $4.27$  & $\textbf{4.15}$ & $\textit{0.887}$\\ 
    $400$   & $5.69$ & $4.78$ & $4.04$  & $3.01$  & $3.00$  & $\textbf{2.91}$ & $\textit{0.593}$\\
    $800$   & $4.58$ & $3.61$ & $2.97$  & $2.11$  & $2.12$  & $\textbf{2.05}$ & $\textit{0.372}$\\
    $1600$  & $3.71$ & $2.74$ & $2.22$  & $1.50$  & $1.51$  & $\textbf{1.46}$ & $\textit{0.265}$\\
    $3200$  & $3.34$ & $2.14$ & $1.67$  & $1.09$  & $1.08$  & $\textbf{1.04}$ & $\textit{0.187}$\\
    $6400$  & $2.76$ & $1.67$ & $1.28$  & $0.791$ & $0.775$ & $\textbf{0.743}$ & $\textit{0.164}$\\ 
    $12800$ & $2.34$ & $1.34$ & $0.998$ & $0.590$ & $0.564$ & $\textbf{0.534}$ & $\textit{0.165}$\\
\end{tabular}
        \label{table:fBm-dev}
        }
    \end{table}
    
    Similarly, Tables \ref{table:arfima-bias} and \ref{table:arfima-dev} provide additional information about the ARFIMA $d$-estimators which appear in Table \ref{table:ARFIMA-MSE} and Figure \ref{fig:ARFIMA-bias_dev} in Section \ref{sec:ARFIMA_results}.
    
    \begin{table}[H]
        \vspace*{0.5cm}
        \centering
        \caption{Absolute area under $d$--bias curve of different $\ARFIMA(0,d,0)$ $d$-estimators by sequence length.}
        \centering
        \resizebox{0.48\textwidth}{!}{
        \begin{tabular}{lccccc}
    & \multicolumn{5}{c}{$\hat{b}_{0.025}$ ($\times 10^{-3}$)}  \\
    \cmidrule(r{4pt}){2-6}
    seq. len. & R/S & variogram & Higuchi & Whittle & $M_{\text{LSTM}}$\\ 
    \midrule
    $100$   & $122$  & $77.5$ & $71.9$  & $38.3$  & $\textbf{14.6}$\\ 
    $200$   & $109$  & $68.8$ & $57.8$  & $19.8$  & $\textbf{7.37}$\\ 
    $400$   & $98.1$ & $63.2$ & $50.9$  & $10.3$  & $\textbf{3.95}$\\ 
    $800$   & $90.1$ & $58.9$ & $48.7$  & $5.48$  & $\textbf{2.11}$\\ 
    $1600$  & $82.7$ & $57.8$ & $46.9$  & $2.70$  & $\textbf{1.17}$\\ 
    $3200$  & $73.3$ & $56.9$ & $45.1$  & $1.60$  & $\textbf{0.614}$\\ 
    $6400$  & $68.3$ & $56.9$ & $44.5$  & $0.862$ & $\textbf{0.350}$\\ 
    $12800$ & $63.6$ & $57.4$ & $43.9$  & $0.495$ & $\textbf{0.216}$\\ 
\end{tabular}
        \label{table:arfima-bias}
        }
    \end{table}
    
    \begin{table}[H]
        \centering
        \caption{Absolute area under $d$--deviation curve of different $\ARFIMA(0,d,0)$ $d$-estimators by sequence length.}
        \centering
        \resizebox{0.48\textwidth}{!}{
        \begin{tabular}{lccccc}
    & \multicolumn{5}{c}{$\hat{\sigma}_{0.025}$ ($\times 10^{-2}$)}  \\
    \cmidrule(r{4pt}){2-6}
    seq. len. & R/S & variogram & Higuchi & Whittle & $M_{\text{LSTM}}$\\ 
    \midrule
    $100$   & $9.84$ & $8.67$ & $8.79$  & $8.85$  & $\textbf{7.21}$\\ 
    $200$   & $7.49$ & $6.48$ & $5.94$  & $5.96$  & $\textbf{5.27}$\\  
    $400$   & $5.82$ & $4.84$ & $4.20$  & $4.12$  & $\textbf{3.81}$\\ 
    $800$   & $4.68$ & $3.64$ & $3.04$  & $2.85$  & $\textbf{2.74}$\\ 
    $1600$  & $3.83$ & $2.77$ & $2.22$  & $\textbf{1.98}$  & $\textbf{1.98}$\\ 
    $3200$  & $3.44$ & $2.18$ & $1.66$  & $\textbf{1.40}$  & $1.42$\\ 
    $6400$  & $2.84$ & $1.72$ & $1.27$  & $\textbf{0.976}$ & $1.03$\\ 
    $12800$ & $2.37$ & $1.40$ & $0.980$ & $\textbf{0.696}$ & $0.744$\\ 
\end{tabular}
        \label{table:arfima-dev}
        }
    \end{table}

\section{Additional information on process generators}\label{sec_stat_tests}
    We implemented process generators which can be used for the simulation of sample paths of iso-normal stochastic processes, including fractional integrals such as fBm and the fOU.
    Focusing on the case of fBm, which is a Gaussian process, standard methodologies for generating Gaussian trajectories can be employed.
    Specifically, for a given sample path length, it suffices to compute a square root decomposition of its covariance matrix evaluated at the specified time points.
    One option is to use Cholesky decomposition, which is applicable in more general settings.
    Alternatively, the circulant embedding method \citep{davies1987tests}, combined with the Fast Fourier Transform (FFT) \citep{kroese2014spatial,dieker2003spectral}, can be utilized.
    To the best of our knowledge, this approach generates sample paths more efficiently than all other known exact methods.    
    
    Let us consider the covariance function of the fBm:
    $$
    \Cov(W_t, W_s) = \frac{1}{2}\left({|t|^{2H} + |s|^{2H} -|t-s|^{2H} }\right), \quad t,s \geq 0,
    $$
    where $H \in (0, 1)$ is the Hurst parameter.
    To generate sample paths of fBm on an equidistant time grid $t_k = k \Delta t$ with $k = 0, 1, \dots, n$, (where $\Delta t>0$ is the fixed distance between discrete time steps e.g. $1/n$), we will consider its increments, that is $X_k = W_{t_{k+1}} - W_{t_{k}}$, which have a variance of $\Var(X_k) = (\Delta t)^{2H}$.
    By considering $Y_k = X_k/(\Delta t)^H$, we obtain a process with unit variance, known as the fractional Gaussian noise (fGn).
    The sequence $\{Y_k\}_{k\in\mathbb{Z}}$ forms a discrete zero-mean Gaussian process with the following covariance function for $k\in\mathbb{Z}$:
    $$
    \Cov(Y_0, Y_{k}) = \frac{1}{2}\left({|k+1|^{2H} - 2|k|^{2H} + |k-1|^{2H}}\right).
    $$
    
    Since the fGn is stationary, its covariance matrix is Toeplitz; thus the sequence $\{Y_k\}$ can be generated efficiently using the circulant embedding approach described in \citep{kroese2014spatial}.
    This method computes the spectrum of the covariance using FFT, multiplies it element-wise by a complex-valued standard normal random vector, and applies the inverse FFT to obtain two independent fGn realizations, significantly reducing computational complexity for large $n$.
    
    From the generated fGn realization $\{Y_k\}$, we can construct an fBm sample path by scaling and cumulative summation:
    $$
    W_{t_k}=(\Delta t)^H\sum_{j=0}^{k-1} Y_j.
    $$

\subsection{Generator speed evaluation}
    Table \ref{tab:process_gen} presents a comparison of execution times for generating fBm sequences using the Python library \texttt{fbm} \citep{pypifbm} and our implementations.
    Among the evaluated approaches, our implementation of Kroese’s method achieved the shortest runtime for generating a single sequence, making it particularly suitable for training our Hurst-estimators, where a large number of realizations with varying Hurst parameters is necessary to build robust models.

    For scenarios that require multiple sequences with the same Hurst parameter, the generator can cache the covariance structure, reducing the computational cost associated with its repeated calculation.
    Although this caching introduces an overhead during the initial sequence generation (resulting in a longer execution time for the first sequence), it significantly reduces the cumulative runtime for subsequent realizations, as shown in the table.
    The covariance structure can also be computed using Cholesky decomposition, which, although substantially slower, benefits similarly from caching.
    The comparative running times for these methods are summarized in the table.

    \begin{table}[H]
        \vspace*{0.23cm}
        \caption{Running times of generating fBm sequences of length $\num{100 000}$.
        }
        \centering
        \begin{tabular}{lccc}
            & & \multicolumn{2}{c}{time}  \\
            \cmidrule(r{4pt}){3-4}
            Method & implementation & 1 seq. & 100 seq.\\
            \midrule
            Cholesky & \texttt{fbm} & $21.5\text{s}$ & $22.8\text{s}$ \\
            Cholesky & Ours & $5.8\text{s}$ & $7.14\text{s}$ \\
            FFT & \texttt{fbm} & $48.4\text{ms}$ & $2.99\text{s}$ \\
            Kroese & Ours (cached)  & $17.7\text{ms}$ & $\textbf{224\text{ms}}$ \\
            Kroese & Ours (on-line) & $\textbf{5.8\text{ms}}$ & $547\text{ms}$ \\
        \end{tabular}
        \label{tab:process_gen}
    \end{table}

  \subsection{Testing with the \texttt{YUIMA} package}\label{sec_yuima}
    To ensure that our results were not biased by dependencies within our project,
    we validated our models on process realizations generated by the R package \texttt{YUIMA} \cite{brouste2014yuima}. 
    The outcomes of the evaluation using neural fBm and fOU Hurst-estimators are presented in Tables \ref{fbm_hurst_yuima} and  \ref{fou_hurst_yuima}, respectively.
    In both instances, the models were trained on process realizations that were generated by our implementation, with the training procedure involving fine-tuning on progressively longer sequences up to the maximum sequence lengths specified in the corresponding tables ($\num{12800}$ and $\num{6400}$).
    Each metric was measured using these final fine-tuned models, which may result in suboptimal performance for shorter sequences when compared to the results presented in the main article.
    The values reported in the tables were computed based on $\num{100 000}$ independent process realizations generated by the \texttt{YUIMA} package.
    For the fOU process, the prior distributions of the parameters were defined according to Section \ref{subsec:fOU_results}. 
    Overall, the models performed similarly to what was observed with our generators, 
    confirming that the models perform as expected even on process realizations generated independently of our implementation.

\section{Additional plots} \label{realizations}
\begin{figure}[b]
        \centering
        \includegraphics[width=0.48\textwidth]{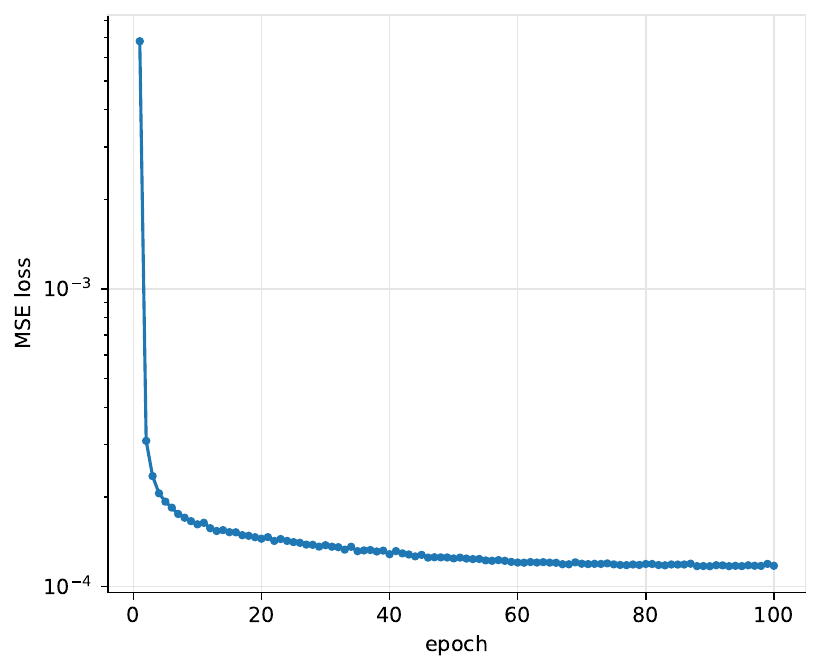}
        \caption{Log-plot of MSE losses of the $M_{\text{LSTM}}$ fBm Hurst-estimator after each training-epoch ($\num{100000}$ sequences). The model was trained and evaluated on sequences of length $n=3200$.}
        \label{fig:training}
        \vspace*{6.52mm}
    \end{figure}

\begin{figure*}[b]
        \centering
        \hspace{-0.2cm}
        \includegraphics[width=0.98\textwidth]{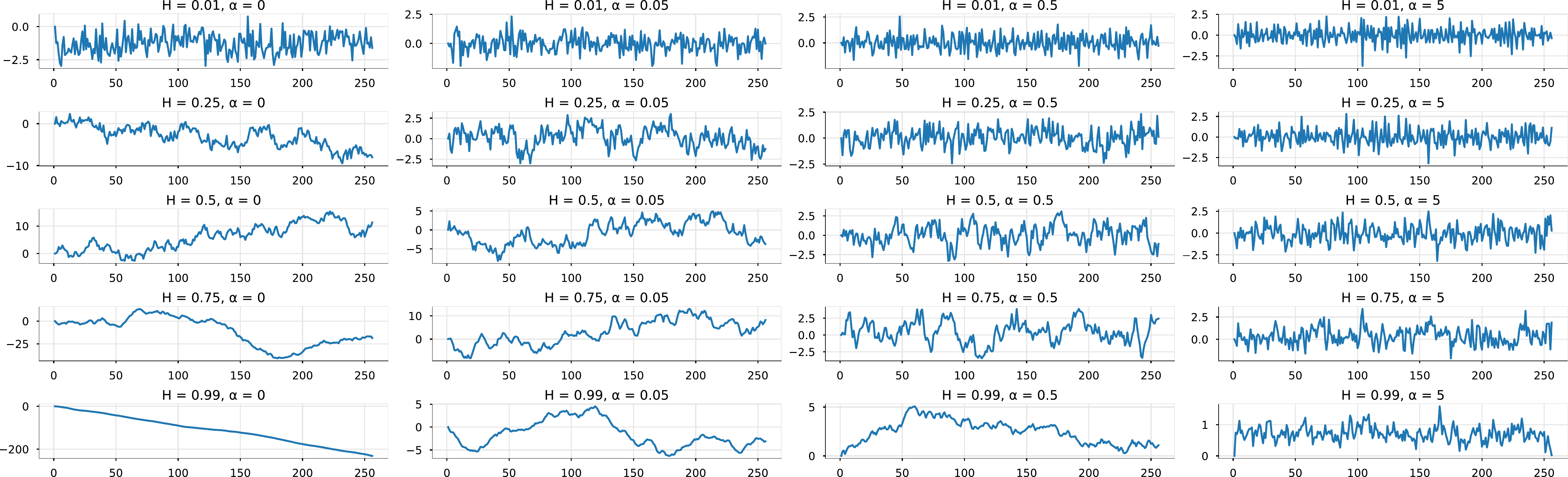}
        \vskip 0.3cm
        \caption{Different fOU realizations by $H$ and $\alpha$.}
        \vskip 0.3cm
        \label{fOU_realizations}
    \end{figure*}

\begin{figure}[t!]
        \centering
        \hspace{-0.35cm}
        \includegraphics[width=0.46\textwidth]{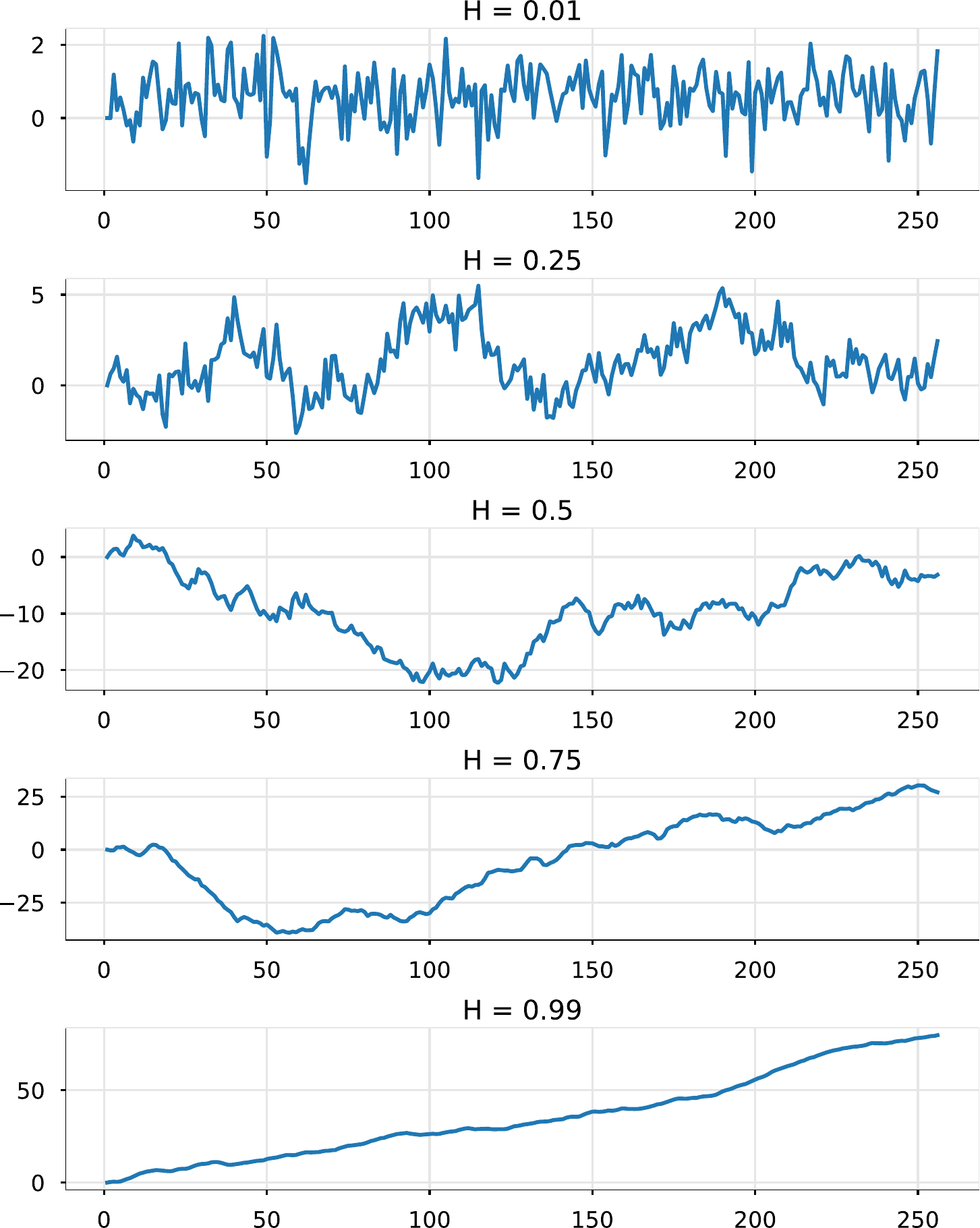}
        \vskip 0.3cm
        \caption{Different fBm realizations by $H$.}
        \label{fBm_realizations}
        \vskip 0.7cm
    \end{figure}

\begin{figure}[t!]
        \centering
        \hspace{-0.3cm}
        \includegraphics[width=0.46\textwidth]{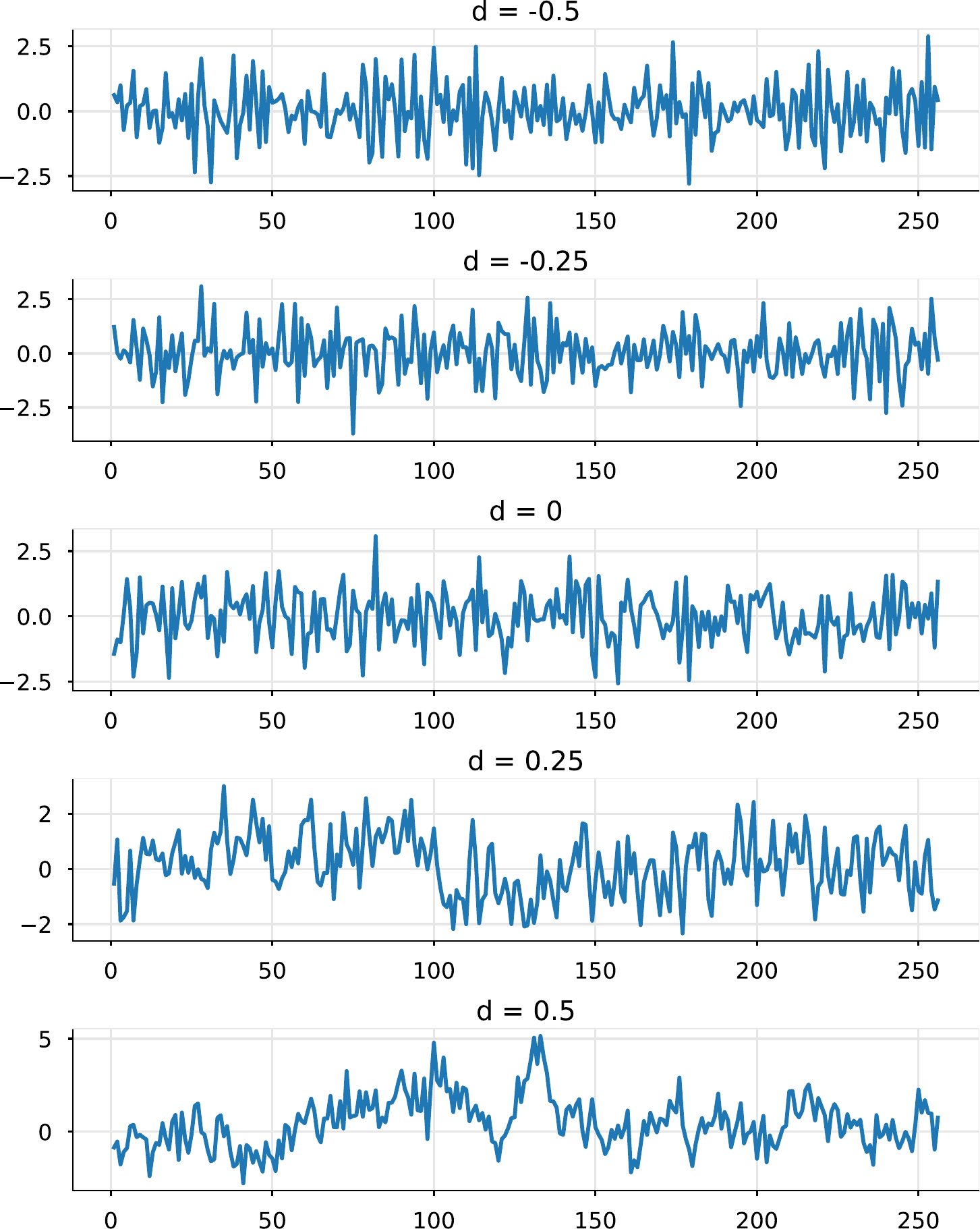}
        \vskip 0.3cm
        \caption{Different $\ARFIMA(0,d,0)$ realizations by $d$.}
        \label{ARFIMA_realizations}
        \vskip 0.7cm
    \end{figure}

    To provide a clearer understanding of the training process and its objectives, in this section we present supplementary visualizations.
    Figure \ref{fig:training} illustrates the typical progression of model loss over the course of a training session, 
    while Figures \ref{fOU_realizations}, \ref{fBm_realizations}, and \ref{ARFIMA_realizations} show different realizations of fOU, fBm and ARFIMA processes by the key target parameters.
    
     \begin{table}[t!]
        \caption{Performance metrics of fBm Hurst-estimators by sequence length on $\num{100 000}$ sequences generated by the \texttt{YUIMA} package.\newline}
        \centering
        \resizebox{0.48\textwidth}{!}{
        \begin{tabular}{lcccccc}
            & \multicolumn{2}{c}{$MSE$ loss ($\times 10^{-3}$)}  & \multicolumn{2}{c}{$\hat{b}_{0.025}$ ($\times 10^{-3}$)} & \multicolumn{2}{c}{$\hat{\sigma}_{0.025}$ ($\times 10^{-2}$)} \\
            \cmidrule(r{4pt}){2-3} \cmidrule(r{4pt}){4-5} \cmidrule(r{4pt}){6-7}
            $\stackrel{\text{seq.}}{\text{len.}}$ & $M_{\text{conv}}$ & $M_{\text{LSTM}}$ & $M_{\text{conv}}$ & $M_{\text{LSTM}}$ & $M_{\text{conv}}$ & $M_{\text{LSTM}}$\\ 
            \midrule
            $100$   & $5.16$  & $4.78$  & $19.3$ & $25.7$ & $6.61$  & $6.12$ \\ 
            $200$   & $2.19$  & $2.07$  & $9.73$ & $12.9$ & $4.41$  & $4.21$ \\ 
            $400$   & $1.00$  & $0.947$ & $4.79$ & $6.10$ & $3.03$  & $2.92$ \\ 
            $800$   & $0.470$ & $0.449$ & $2.47$ & $3.33$ & $2.10$  & $2.03$ \\ 
            $1600$  & $0.236$ & $0.222$ & $1.84$ & $2.04$ & $1.49$  & $1.43$ \\ 
            $3200$  & $0.121$ & $0.111$ & $1.88$ & $1.46$ & $1.06$  & $1.011$ \\  
            $6400$  & $0.067$ & $0.057$ & $2.68$ & $1.35$ & $0.748$ & $0.715$ \\ 
            $12800$ & $0.041$ & $0.030$ & $3.07$ & $1.40$ & $0.535$ & $0.511$ \\  
        \end{tabular}
        }
        \label{fbm_hurst_yuima}
    \end{table}   

    \begin{table}[t!]
        \caption{Performance metrics of $M_{\text{LSTM}}$ the fOU Hurst-estimator by sequence length, measured on $\num{100 000}$ sequences generated by the \texttt{YUIMA} package.\newline}
        \centering
        \resizebox{0.48\textwidth}{!}{
        \begin{tabular}{lccc}
            seq. len. & \multicolumn{1}{c}{MSE ($\times 10^{-3}$)} & \multicolumn{1}{c}{$\hat{b}_{0.025}$ ($\times 10^{-3}$)} & \multicolumn{1}{c}{$\hat{\sigma}_{0.025}$ ($\times 10^{-2}$)}\\
            \midrule
            $100$ & $8.19$ & $16.0$ & $8.37$ \\ 
            $200$ & $2.85$ & $7.23$ & $5.08$ \\ 
            $400$ & $1.22$ & $4.13$ & $3.35$\\ 
            $800$ & $0.577$ & $2.79$ & $2.32$ \\ 
            $1600$ & $0.293$ & $2.19$ & $1.65$ \\ 
            $3200$ & $0.151$ & $1.94$ & $1.17$ \\ 
            $6400$ & $0.081$ & $2.25$ & $0.839$ \\ 
        \end{tabular}
        }
        \label{fou_hurst_yuima}
        \vspace*{3.5mm}
    \end{table}

\end{document}